\documentclass[letterpaper, 10 pt, conference]{ieeeconf}  

\IEEEoverridecommandlockouts                              

\usepackage{graphics} 
\usepackage{epsfig} 
\usepackage{mathptmx} 
\usepackage{times} 
\usepackage{amsmath} 
\usepackage{amssymb}  

\usepackage{microtype}
\usepackage{subfigure}
\usepackage{booktabs} 
\usepackage{xfrac}
\usepackage{nicefrac}
\usepackage{multirow}
\usepackage{multicol}
\usepackage{hyperref}
\usepackage{wrapfig}
\usepackage{xcolor}
\usepackage{array}
\usepackage{algorithm}
\usepackage{gensymb}
\usepackage{xr}

\title{\LARGE \bf
Learning-Driven Exploration for Reinforcement Learning
}

\author{Muhammad Usama and Dong Eui Chang*
\thanks{$^{1}$The authors are with Control Laboratory, School of Electrical Engineering,
        Korea Advanced Institute of Science and Technology (KAIST), Daejeon, Republic of Korea.
        {\tt\small \{usama, dechang\}@kaist.ac.kr}}%
\thanks{* corresponding author}
}

\begin{document}

\maketitle
\thispagestyle{empty}
\pagestyle{empty}

\begin{abstract}

Effective and intelligent exploration has been an unresolved problem for reinforcement learning. Most contemporary reinforcement learning relies on simple heuristic strategies such as $\epsilon$-greedy exploration or adding Gaussian noise to actions. These heuristics, however, are unable to intelligently distinguish the well explored and the unexplored regions of  state space, which can lead to inefficient use of training time. We introduce entropy-based exploration (EBE) that enables an agent to explore efficiently the unexplored regions of  state space. EBE quantifies the agent's learning in a state using merely state-dependent action values and adaptively explores the state space, i.e. more exploration for the unexplored region of the state space. We perform experiments on a diverse set of environments and demonstrate that EBE enables efficient exploration that ultimately results in faster learning without having to tune any hyperparameter. 

The code to reproduce the experiments is given at \url{https://github.com/Usama1002/EBE-Exploration} and the supplementary video is given at \url{https://youtu.be/nJggIjjzKic}.


\end{abstract}

\section{Introduction}

Reinforcement learning (RL) is a sub-field of machine learning where an agent interacts with its environment to learn a policy that maximizes the cumulative reward over a horizon. Since the agent does not begin with perfect knowledge of the environment dynamics, it has to learn solving the task through the process of exploration, thus, giving rise to the fundamental trade-off between exploration vs exploitation. A long-standing problem in RL is to find ways to achieve better trade-off between exploration and exploitation.

\begin{figure*}[t]
    \centering
    \includegraphics[scale=0.3]{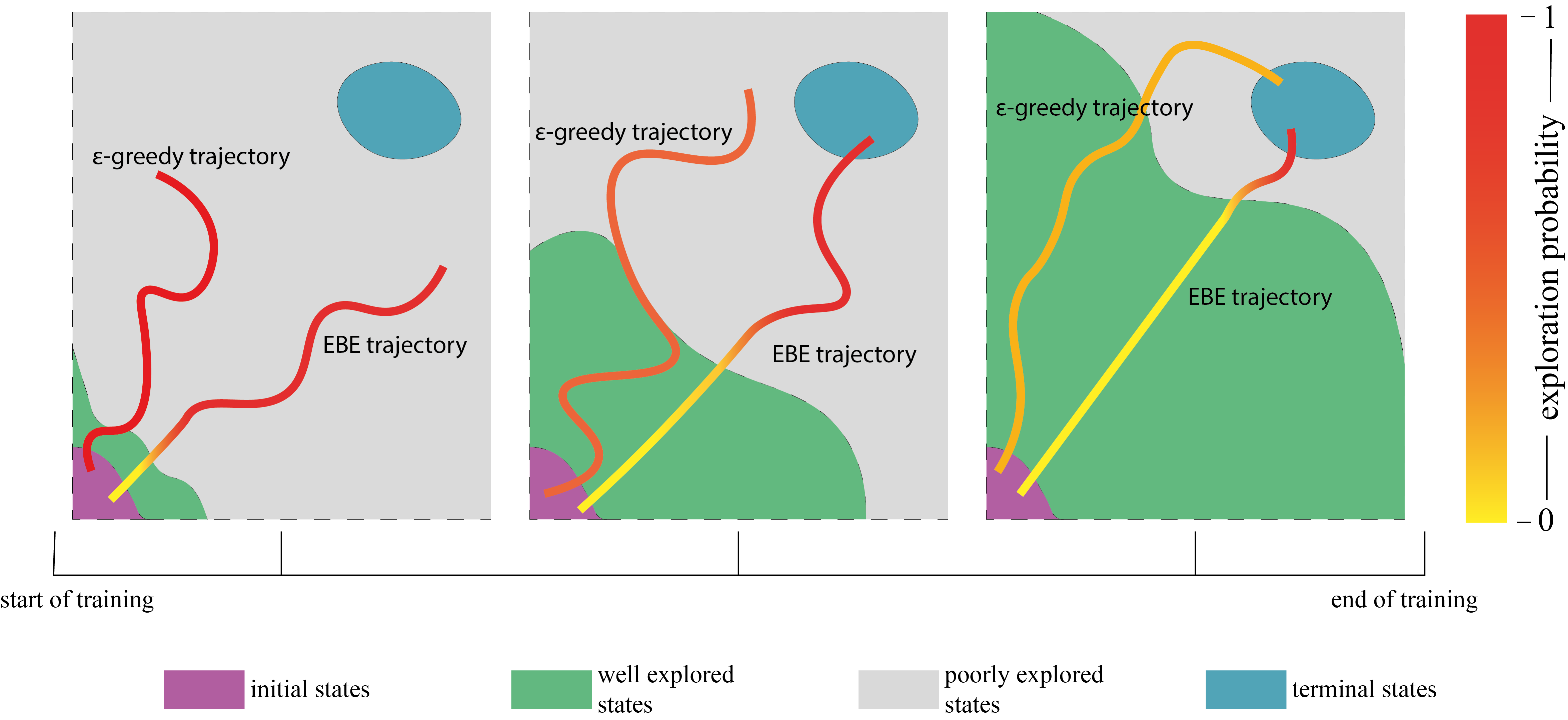}
    \caption{Conceptual visualization of entropy-based exploration (EBE).}
    \label{fig:concept}
\end{figure*}

In this work, we argue that state dependent action values can provide valuable information to the agent about its learning progress in a state. We use the concept of entropy from information theory to quantify agent's learning in a state and the algorithm subsequently decides whether to explore in a state based on it. This minimizes the prospects of unnecessary exploration while still exploring the poorly explored regions of the state space.


\section{Preliminaries}
\subsection{Reinforcement Learning}
Reinforcement learning \cite{suttonAndBarto} is a sequential decision making process in which an agent, while in state $s_{t} \in \mathcal{S}$  of environment $\mathcal{E}$, at time step $t$, chooses an action $a_{t}$ from set $\mathcal{A}$ following a policy $\pi(s)$ and receives a reward $r_{t}$ and $\mathcal{E}$ transitions into next state $s_{t+1}$ following transition or dynamics model $\mathcal{P}$. The goal of any RL algorithm is to maximize the expected discounted return $R_t = \mathbb{E}_{\pi, \mathcal{P}}[\sum_{\tau=t}^{\infty}\gamma^{\tau-t}r_{\tau}]$, where $\gamma$ is the discount factor. Given a policy $\pi$, the state dependent action and the state value functions are defined as $Q^{\pi}(s,a) = \mathbb{E}[R_t | s_t = s, a_t = a, \pi ]$ and $ V^{\pi}(s) = \mathbb{E}_{a \sim \pi(s)}[Q^{\pi}(s,a)]$, respectively. We use a deep $Q$-network (DQN) \cite{humanlevel} to approximate the high-dimensional action value function, $Q^{\pi}$.

\subsection{Entropy}
The entropy $H_X$ of a discrete random variable $X$ with probability distribution $p_{X}(x)$ is defined as
\begin{align*}
H_X & = - \sum_{x \in \mathcal{X}} p_{X}(x) \log_{b}p_{X}(x) \\
    & = - \displaystyle \mathbb{E}_{X \sim p_{X}} [\log_{b} p_{X}(x)].
\end{align*}

\section{Entropy-Based Exploration (EBE)}
In this section, first, we go through the motivation behind EBE and then we present the mathematical realization of the concept.

\subsection{Motivation}
\label{section:deep_exploration}
An efficient exploration strategy should adapt itself to explore more in poorly explored regions of the state space, which we refer to as \textit{learning-driven exploration}. This allows the agent to explore deeper into the state space resulting in \textit{deep\footnote{the word \textit{deep} is used here in different context from \textit{deep learning}.} exploration}. Our definition of deep exploration is different from \cite{osband_deep_exploration} where deep exploration means "exploration which is directed over multiple time steps or far-sighted exploration" \cite{osband_deep_exploration}. In our work, \textit{deep exploration} concerns spatially extended exploration in the state space.

This concept is illustrated in Figure \ref{fig:concept} where EBE and $\epsilon$-greedy exploration are depicted by two separate trajectories at three different stages of the presumed training process. The redness of a trajectory indicates the exploration probability in that state. The exploration probability for EBE increases as we move towards an unexplored region of the state space. But for $\epsilon$-greedy exploration where the value of $\epsilon$ is annealed from the start to the end of the training process, at a particular instant in training, the agent explores in all states with the same probability irrespective of the knowledge it already has acquired. Adaptive exploration by EBE enables the agent to allocate more resources towards exploring the relatively poorly understood regions of the state space, thus improving the learning progress.

\subsection{Entropy-Based Exploration (EBE): A Realization of Learning-Driven Deep Exploration}
The agent quantifies the utility of an action in a state in the form of state-dependent $Q$-values. 
We use the difference between $Q(s,a)$, where $a \in \mathcal{A}$, in a state $s$ as an estimate of the agent's learning progress in that state.
Therefore, we use $Q$-values to define a probability distribution over actions in a state, i.e.
\begin{equation}
    p_s(a) = \frac{e^{Q(s,a)-\max_{\Tilde{a} \in \mathcal{A}} Q(s,\Tilde{a})}}{\sum_{b \in \mathcal{A}}e^{Q(s,b)-\max_{\Tilde{a} \in \mathcal{A}} Q(s,\Tilde{a})}}.
        \label{eq:action_distribution}
\end{equation}
Since $e^{Q(s,a)}$ may cause numerical overflow when ${Q(s,a)}$ is large, we use the so-called \textit{max trick}  in equation \eqref{eq:action_distribution} to improve numerical stability.

We use $p_s(a)$ to obtain the normalized state dependent entropy, $H(s) \in [0,1]$, as follows
\begin{align}
    H(s) = - \sum_{a \in \mathcal{A}} p_{s}(a) \log_{|\mathcal{A}|}p_{s}(a) 
    \label{eq:EBE_base}.
\end{align}

The entropy $H(s)$ in equation \eqref{eq:EBE_base} quantifies the agent's learning in state $s$: the lower the entropy $H(s)$, the more learned the agent is as some actions are better than the others. Therefore, we use $H(s)$ to guide exploration in a state. Given $H(s)$ in a state from equation \eqref{eq:EBE_base}, the agent explores with probability $H(s)$ i.e. it behaves randomly. In practice, EBE is similar to $\epsilon$-greedy exploration method with $\epsilon$ replaced with the state dependent $H(s)$.

\begin{figure}
    \centering
    \subfigure[]
    {
        \includegraphics[scale=.3]{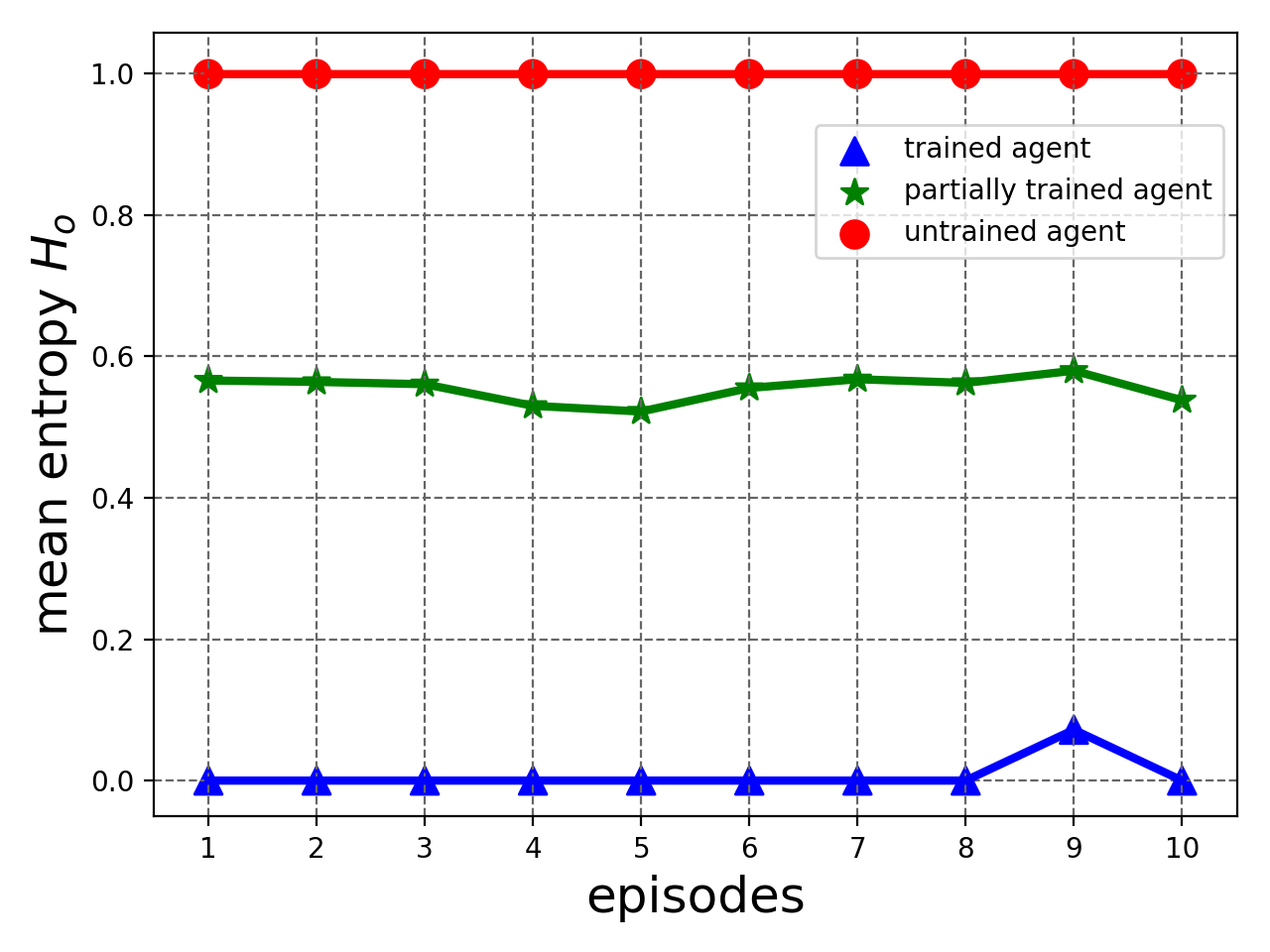}
    }
    \\
    \subfigure[]
    {
        \includegraphics[scale=.3]{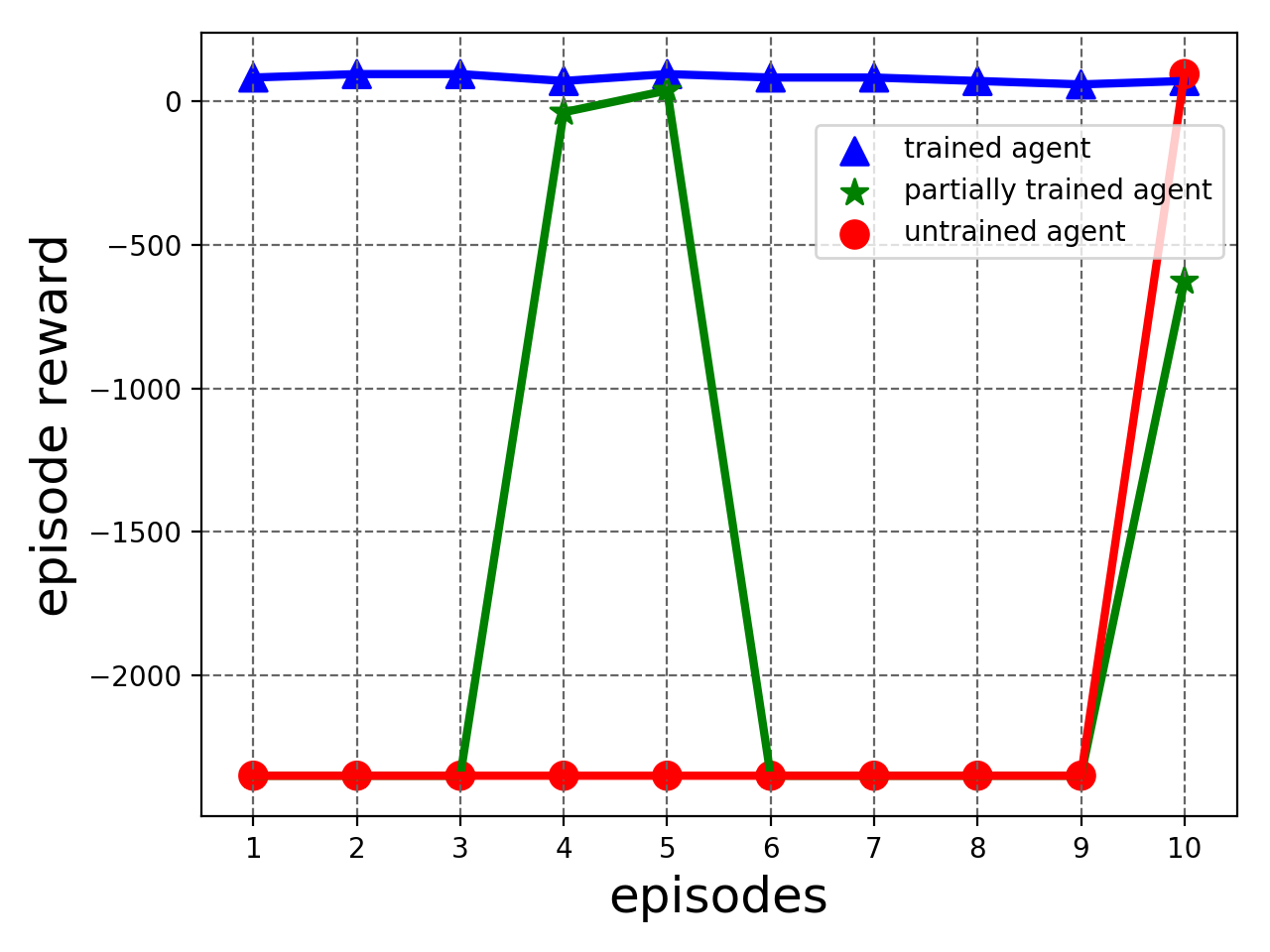}
    }
    \caption
    {
    Plot of (a) mean entropy $H_{o}$, given in equation \eqref{eq:mean_entropy}, and (b) accumulative episode reward for trained, partially trained and untrained agents for 10 test episodes. The agents are trained to play VizDoom game Seek and Destroy.
    }
    \label{fig:concept_entropy_reward}
    \vspace{-12.5pt}
\end{figure}

\subsubsection{How does entropy estimate agent's learning in a state?}
The state space can be broadly classified into two categories: states in which a choice of action is crucial and states in which a choice of action does not significantly impact what happens in the future \cite{dueling_networks}. For the former set of states, some actions are decisively better than the others. Quantitatively, it means that $Q$-values of the better actions are significantly higher than $Q$-values of the remaining actions. Therefore, the distribution defined in equation \eqref{eq:action_distribution} is highly skewed towards better actions and by equation \eqref{eq:EBE_base}, the entropy of these states is low. Note that the lowest achievable entropy may be different for different states.

Consider, for example, the case where the agent is trained to play VizDoom game Seek and Destroy (see Section \ref{subsubsection:seekndestroy} for details). We consider three cases consisting of an untrained agent\footnote{the $Q$-network was initialized using Kaiming Uniform method \cite{kaiming_initialization} and no further training was performed.}, a partially trained agent\footnote{the agent was trained using EBE for two epochs only.} and a trained agent\footnote{the agent was trained using EBE for 20 epochs}. Here, we define $H_{o} \in [0,1]$ as entropy averaged over an entire episode, i.e.
\begin{equation}
    H_{o} = \frac{1}{N} \sum_{i=1}^{N}H(s_i) ,
    \label{eq:mean_entropy}
\end{equation}
where $N$ is the number of steps in the episode, $s_i$ represents the state at $i^{\text{th}}$ step and $H(s_i)$ is given in equation \eqref{eq:EBE_base}. We test the agents for 10 episodes. Figure \ref{fig:concept_entropy_reward} plots $H_{o}$ and the accumulated episode reward versus test episodes. We see that the trained agent has the lowest $H_o$ (Fig. \ref{fig:concept_entropy_reward}(a)) and the highest accumulative reward (Fig. \ref{fig:concept_entropy_reward}(b)) for all episodes. The partially trained agent still has significant $H_{o}$ for all episodes which reflects its incomplete learning while untrained agent has the highest $H_{o}$ and the lowest reward.

These results show that entropy is a good measure to estimate agent's learning in a state, which in turn can be used to quantify the need for exploration. This forms the basis for our proposed entropy-based exploration strategy.

\subsubsection{How EBE is different from Boltzmann Exploration?}
EBE uses entropy of a state $H(s)$, as defined in equation \eqref{eq:EBE_base}, to decide \textit{whether} to explore in a state. Boltzmann exploration, on the other hand, does not use the entropy of a state and probabilistically explores in a state based on probability $p_{Boltzmann}(a) = \nicefrac{e^{Q(s,a)/\tau}}{\sum_{b \in \mathcal{A}}e^{Q(s,b)/\tau}}$, where $\tau$ is the temperature.

\section{Experiments}
\label{section:experiments}
We demonstrate the performance of EBE on many environments including a linear environment, a simpler breakout game and multiple FPS games of Vizdoom \cite{vizdoom}. Results shown are averaged over five runs.

\subsection{Value Iteration on Simple Linear Environment}
\label{subsection:linear}
We start experiments on a simple value iteration task as this task is devoid of many confounding complexities and provides better insight into used methods. Moreover, exact optimal $Q$-values, $Q^{*}(s,a)$ for all $(s,a) \in (\mathcal{S} \times \mathcal{A})$, can be computed analytically which helps monitor the learning progress.

\begin{figure}
    \centering
    \subfigure[]
    {
        \includegraphics[scale=.1]{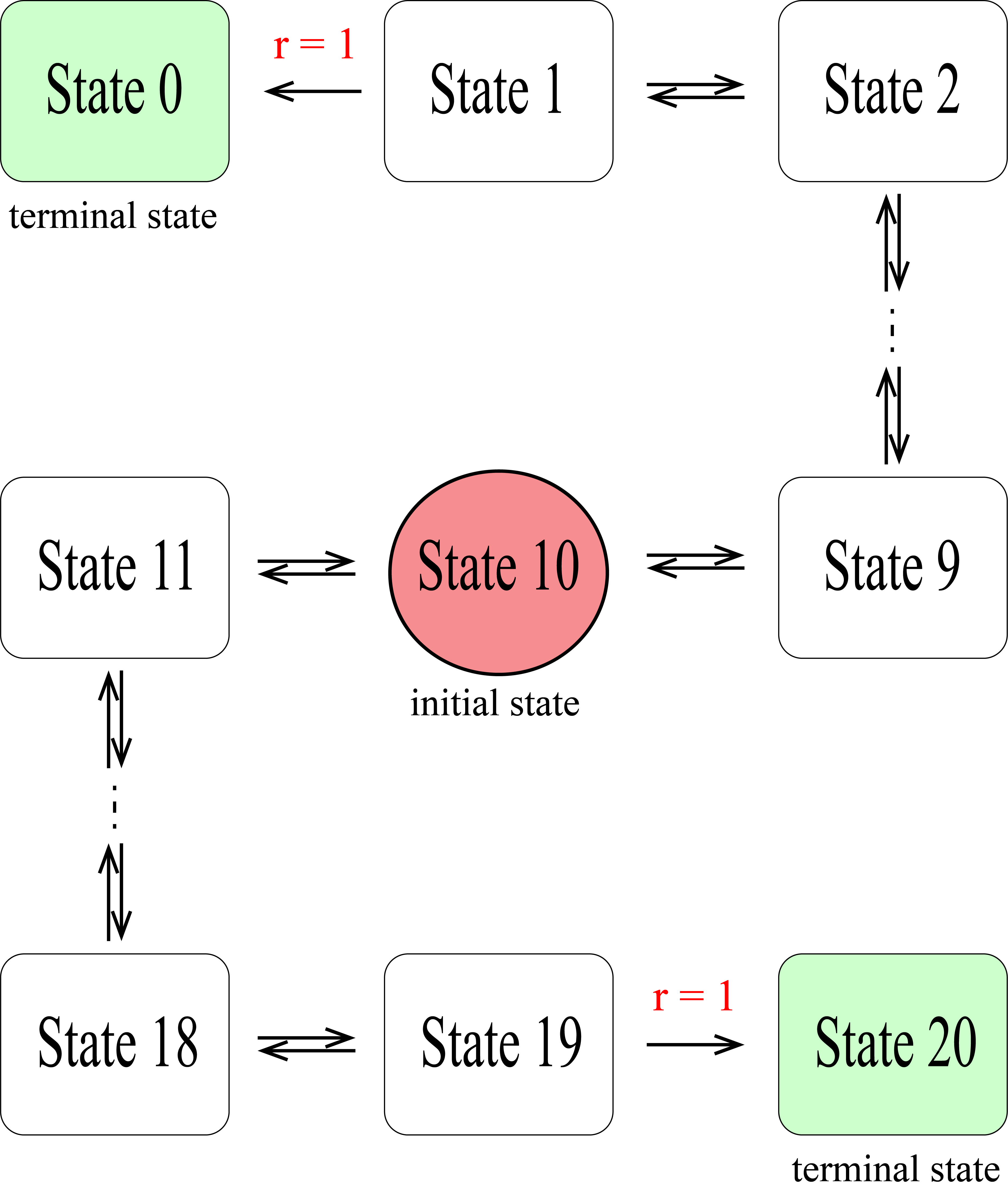}
    }
    \\
    \subfigure[]
    {
        \includegraphics[scale=.3]{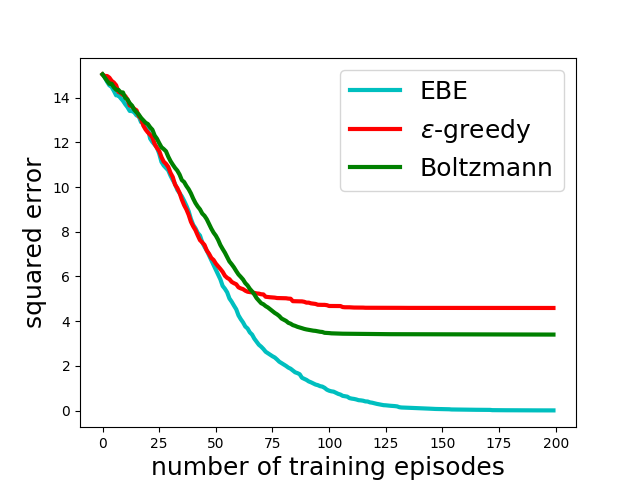}
    }
    \caption
    {
        (a) Simple linear environment consists of 21 states. Episode starts in state $s=10$, shown in red circle. States $s=0$ and $s=20$, shown in green rounded rectangles, are terminal states. For non-terminal states, the agent can transition into either of its neighboring states.
    The agent gets reward $r=1$ for transitioning into the terminal states and zero reward otherwise. (b) Squared Error loss for value iteration task on linear environment.
    }
    \label{fig:linear_environment_and_squared_error}
    \vspace{-12.5pt}
\end{figure}

The environment is described in Figure \ref{fig:linear_environment_and_squared_error}(a). We use temporal difference based tabular Q-learning without eligibility traces to learn the optimal $Q$-values, $Q(s,a)$ for all $(s,a) \in (\mathcal{S} \times \mathcal{A})$.

As baselines, we use $\epsilon$-greedy exploration where the value of $\epsilon$ is linearly annealed from 1.0 to 0.0 over the number of episodes and Boltzmann exploration where the temperature is linearly decreased from 0.8 to 0.1. The evaluation metric is the mean squared error between the actual $Q$-values, $Q^{*}(s,a)$, and the learned $Q$-values, $Q(s,a)$:
\[
\mathcal{L} = \sum_{s \in \mathcal{S}, a \in \mathcal{A}} (Q^{*}(s,a) - Q(s,a))^2.
\]
The squared error is plotted in Figure \ref{fig:linear_environment_and_squared_error}(b). We see that $Q$-values learnt with EBE converge to the optimal $Q$-values while others fail. This is a very promising result as it indicates the ability of EBE to adequately explore the state space.

\begin{figure*}[t]
    \centering
    \subfigure[]
    {
        \includegraphics[scale=.25]{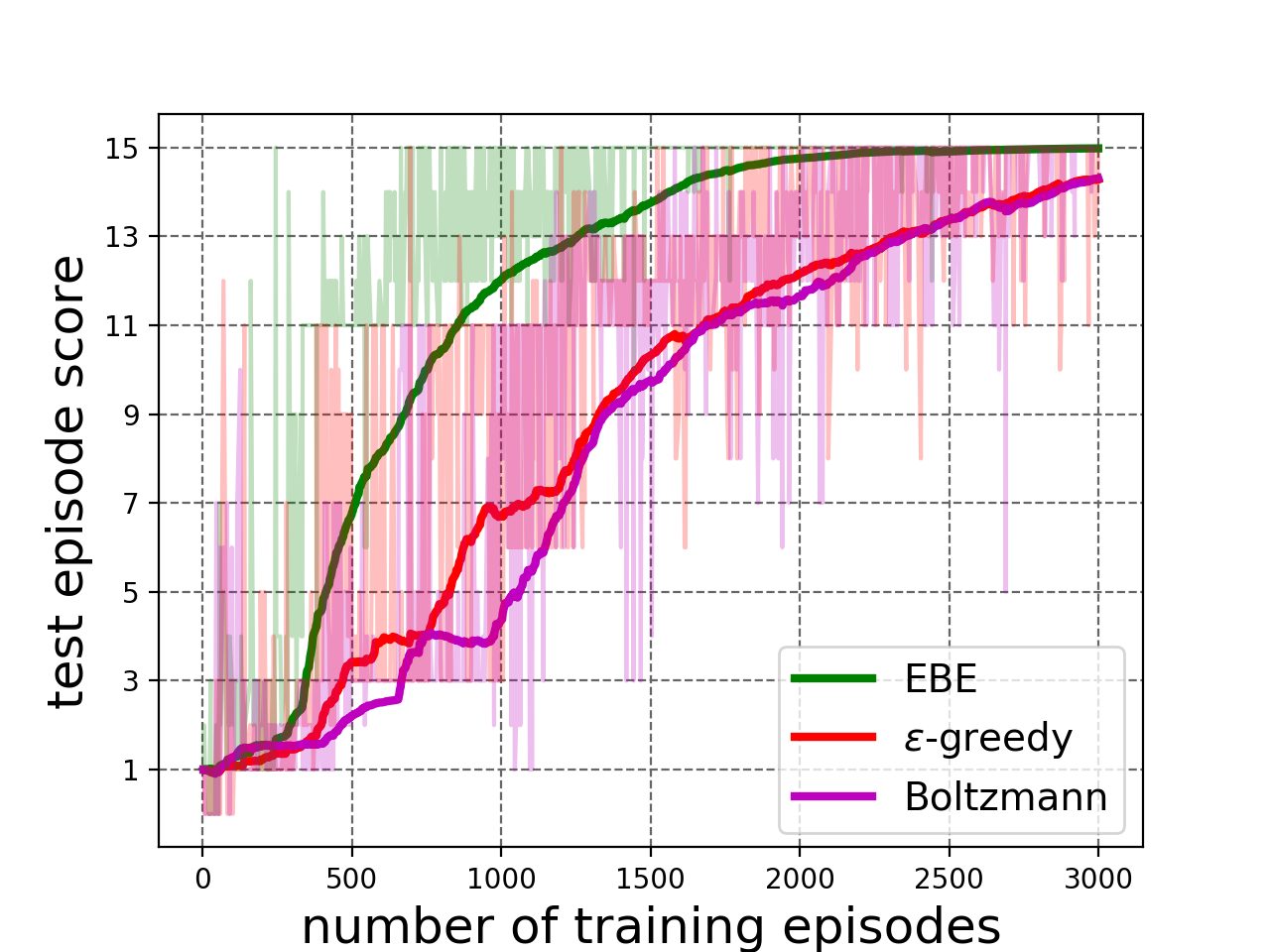}
    }
    \subfigure[]
    {
        \includegraphics[scale=.25]{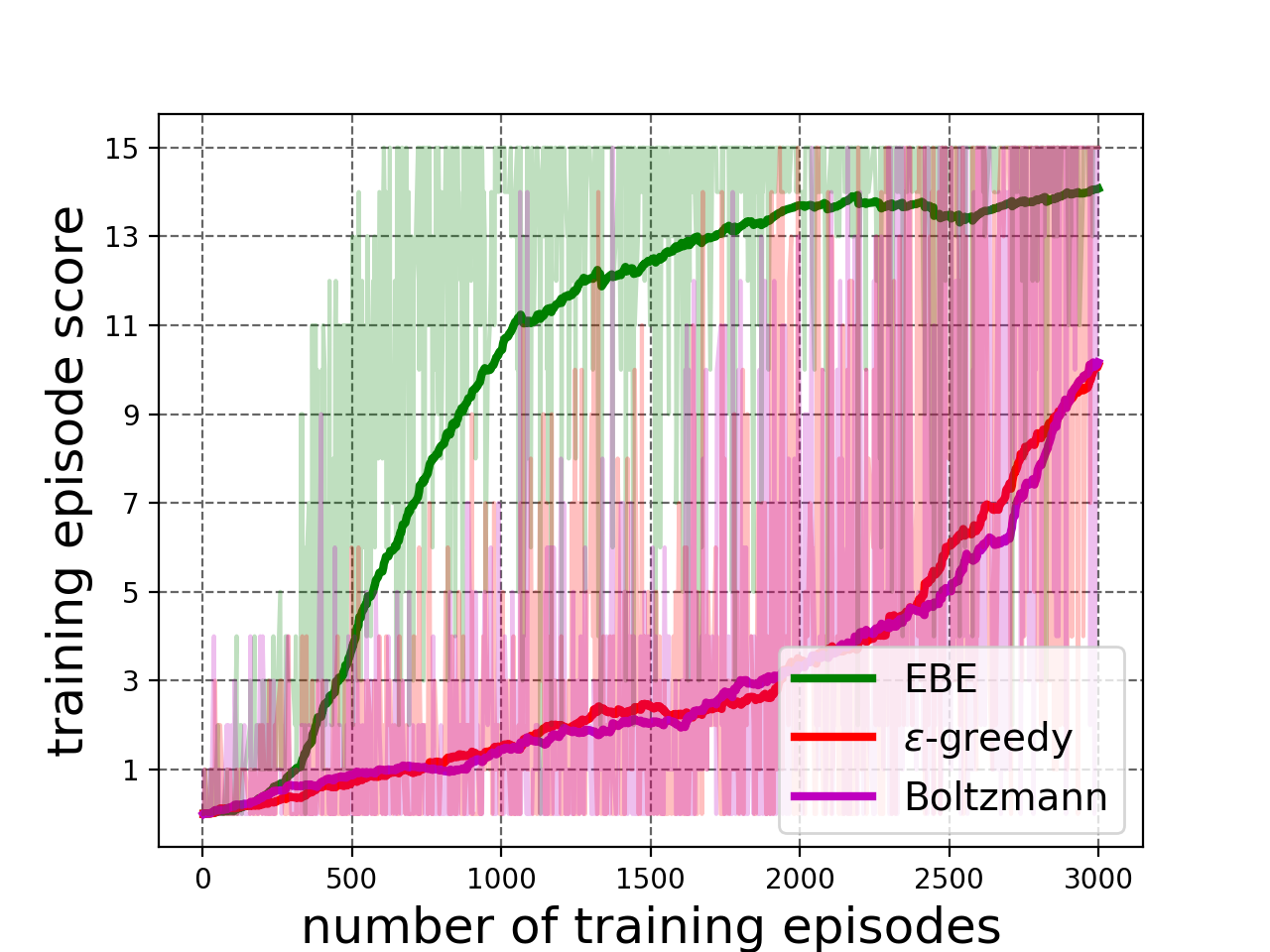}
    }
        \subfigure[]
    {
        \includegraphics[scale=.25]{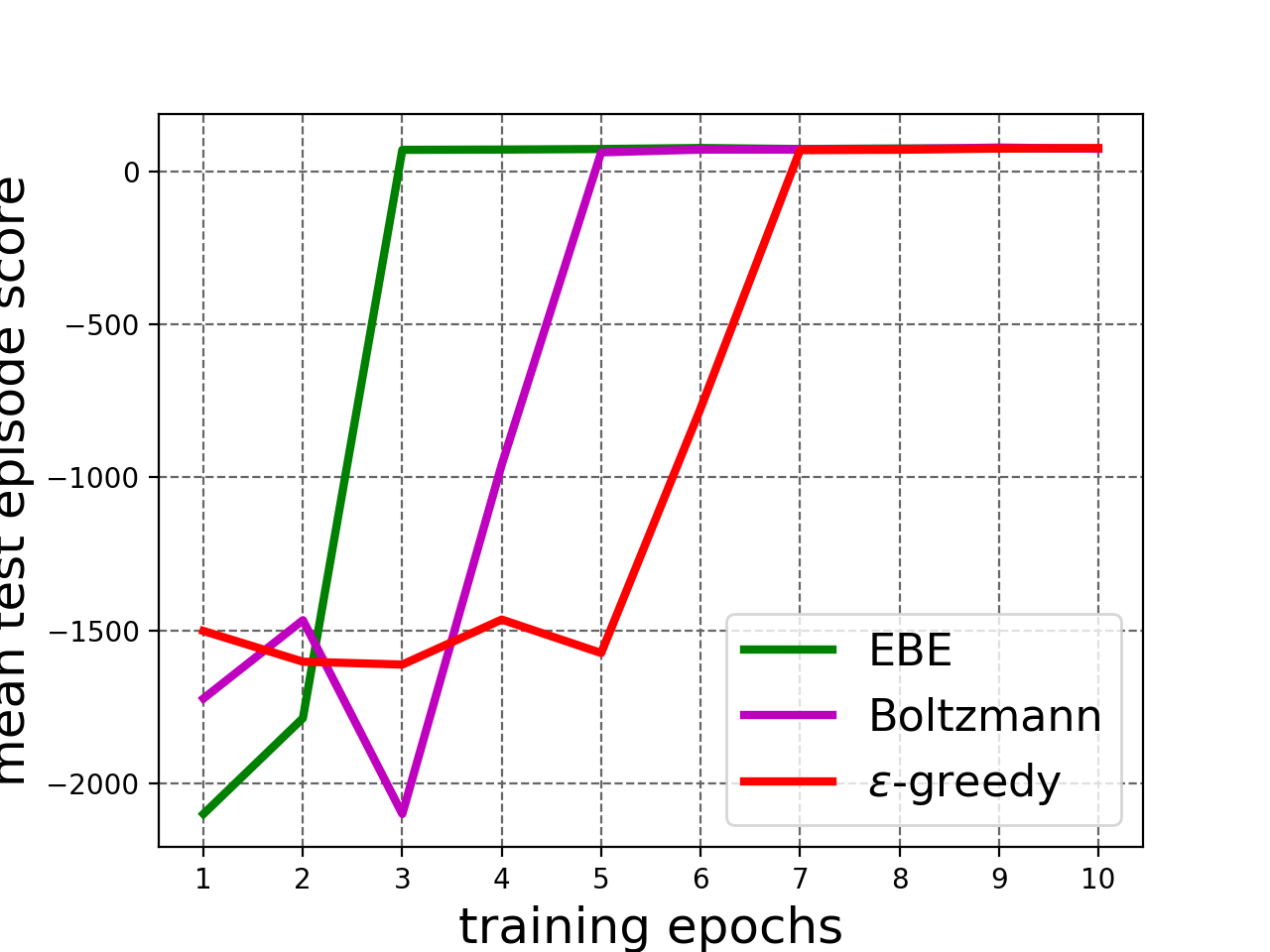}
    }
    \subfigure[]
    {
        \includegraphics[scale=.25]{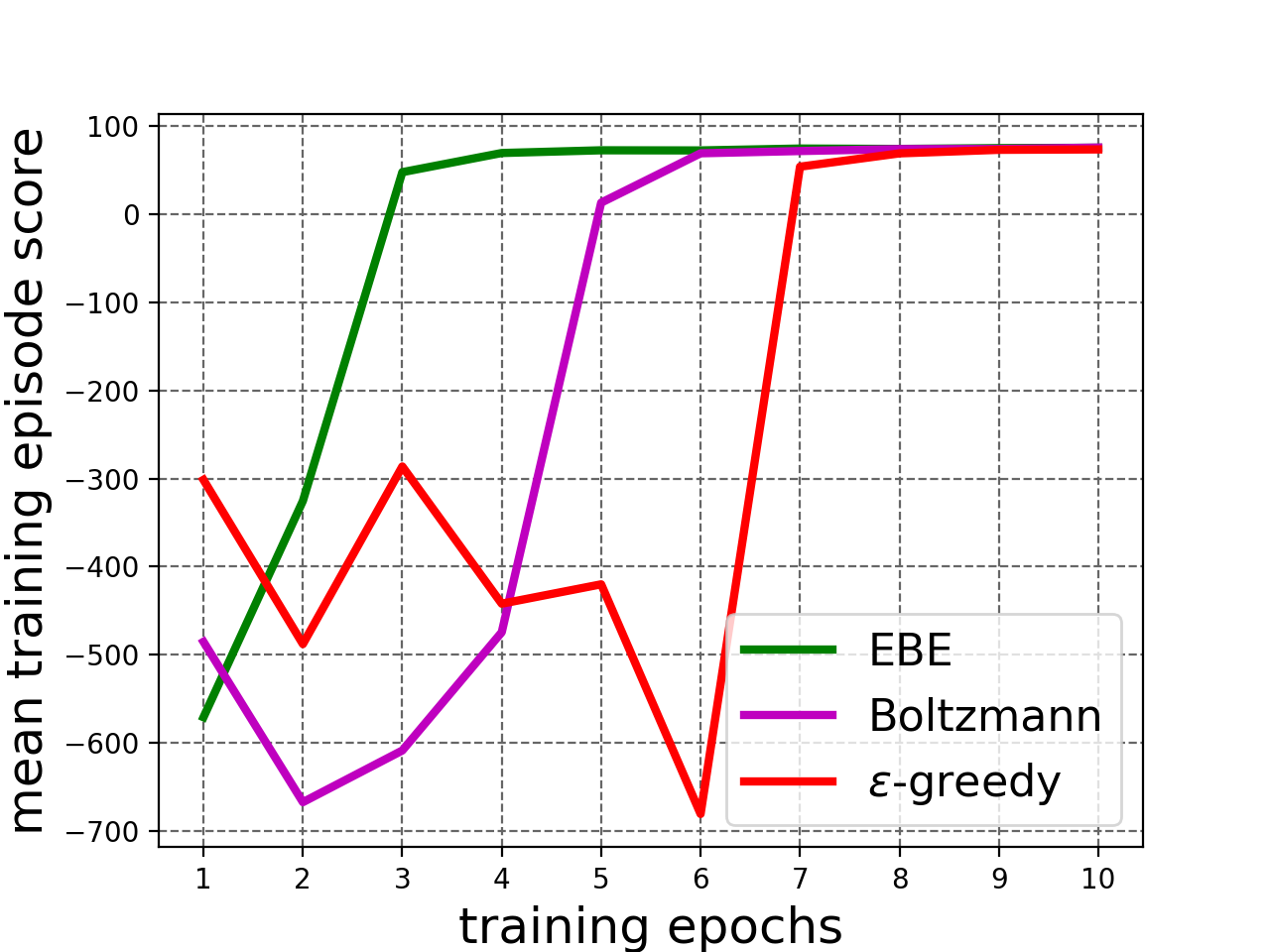}
    }
    \caption
    {
        Plots show (a) test episode scores and (b) training episode scores for agents trained with EBE, $\epsilon$-greedy exploration and Boltzmann exploration on breakout game.
        Likewise, (c) plots mean test score of 100 test episode scores played after each training epoch and (d) plots mean score of all training episodes played in a training epoch for VizDoom game Seek and Destroy.
        Smoothed data is shown with solid lines while unsmoothed data is ghosted in the background. Smoothing method is adopted from \cite{tensorboard} with weight 0.99.
    }
    \label{fig:simpler_breakout_results_seek_and_find_results}
    \vspace{-12.5pt}
\end{figure*}

\subsection{Breakout Game}
Next, we experiment with breakout game complex enough to offer significant learning challenge as it uses a neural network as a function approximator and works on raw images as states. There are 15 bricks in total and the agent gets a reward of 1 point for breaking a brick. An episode ends when one of the following happens: all bricks are broken, the paddle misses the ball or the maximum number of steps has been reached. We use a stack of 2 images: the current image and the previous image, as our state observation. In a state, the agent opts to do nothing or move the paddle left or right. EBE is compared to $\epsilon$-greedy exploration in which $\epsilon$ is linearly annealed from 1.0 to 0.0 over the number of episodes and Boltzmann exploration where temperature is linearly annealed from 1.0 to 0.01 over training process. See Appendix \ref{appendix-breakout} for details regarding the experimental setup.

The results are shown in Figure \ref{fig:simpler_breakout_results_seek_and_find_results}. EBE results in faster learning than baselines (Figure \ref{fig:simpler_breakout_results_seek_and_find_results}(a)). As seen in Figure \ref{fig:simpler_breakout_results_seek_and_find_results}(b), the agent trained with EBE starts performing episodes with higher reward early on in the training process as compared to the other agents, which validates our hypothesis of \textit{deep exploration}, in which the agent transitions quickly into the poorly explored region of the state space, which usually corresponds to the later states of a training episode.

\subsection{VizDoom}
\label{subsection:vizdoom}
We use the following environments from VizDoom platform \cite{vizdoom} to conduct experiments.
\subsubsection{Seek and Destroy}
\label{subsubsection:seekndestroy}
Here, the agent is tasked to shoot an attacking monster spawned randomly on the opposite wall of the room. The gun can only fire straight, so the agent must come in line with the monster before firing. The agent gets a reward of 101 point for shooting the monster, $-5$ for firing each shot and $-1$ for each step taken. The agent gets raw images as state observations. It can either move left, move right or fire a shot in a state. The episode ends when either the monster is dead, the player is dead or 300 time steps have passed.

We compare EBE with Boltzmann and $\epsilon$-greedy exploration strategies. In Boltzmann exploration, the temperature parameter is linearly annealed from 1.0 to 0.01 over the training epochs. For $\epsilon$-greedy exploration, $\epsilon$ is set to 1.0 for first epoch, then $\epsilon$ is linearly annealed to 0.01 till epoch 6. Thereafter, $\epsilon=0.01$ is used. See Appendix \ref{appendix-vizdoom} for details about the training setup.

The results are shown in Figure \ref{fig:simpler_breakout_results_seek_and_find_results}. Mean test scores in Figure \ref{fig:simpler_breakout_results_seek_and_find_results}(c) show that EBE-trained agent outperforms the agents trained with baselines. Similarly, we see in Figure \ref{fig:simpler_breakout_results_seek_and_find_results}(d) that EBE exploration results in high reward training episodes considerably earlier in training manifesting deep exploration (Section \ref{section:deep_exploration}).

\subsubsection{Defend the Center (DTC)}
\label{subsubsection:dtc}

\begin{table}[t]
\caption{Baseline exploration strategies for DTC and DTL experiments}
\label{tab:concept}
\begin{center}
\begin{small}
\begin{tabular}{l c}
\toprule
variant &  \hfil details  \\
\midrule
$\epsilon$-greedy I   &  \parbox{5cm} {$\epsilon=$1.0 is used for first 100 epochs, then it is linearly annealed to 0.01 till 600 epochs. Afterwards $\epsilon=$0.01 is used.} \\ 
\midrule
$\epsilon$-greedy II  &  \parbox{5cm}{$\epsilon$ is linearly annealed from 1.0 to 0.01 over the entire training process.}\\
\midrule
$\epsilon$-greedy III  &  \parbox{5cm}{$\epsilon=1.0$ is used for first 100 epochs. $\epsilon$ is then linearly annealed from 1.0 to 0.01 over the remaining training process.} \\
\midrule
Boltzmann & \parbox{5cm}{temperature is linearly annealed from 1.0 to 0.01.} \\
\bottomrule
\end{tabular}
\end{small}
\end{center}
\vskip -0.35in
\label{tab:variants-of-egreedy}
\end{table}

\begin{figure*}[t]
    \centering
    \subfigure[]
    {
        \includegraphics[scale=.25]{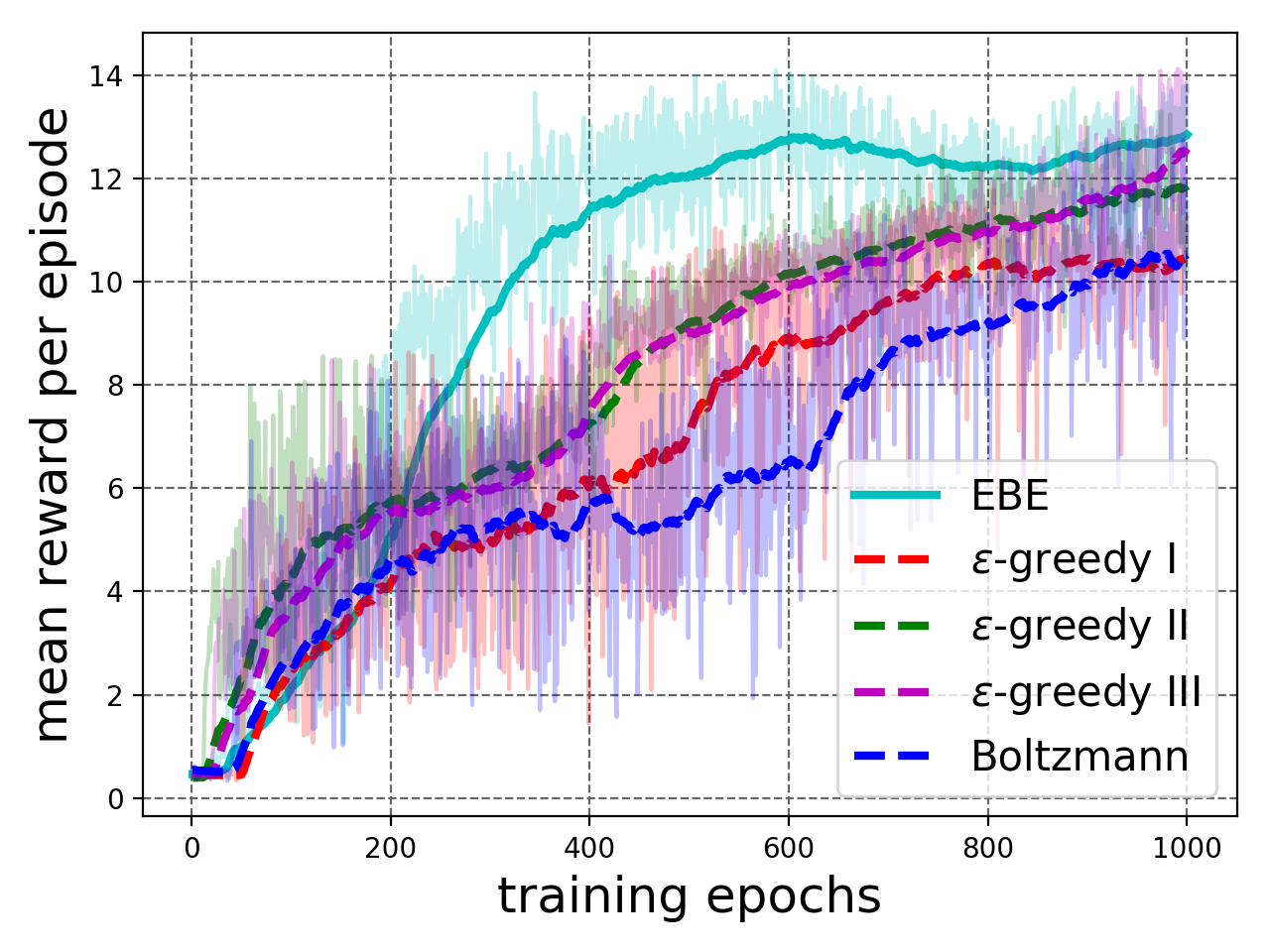}
    }
    \subfigure[]
    {
        \includegraphics[scale=.25]{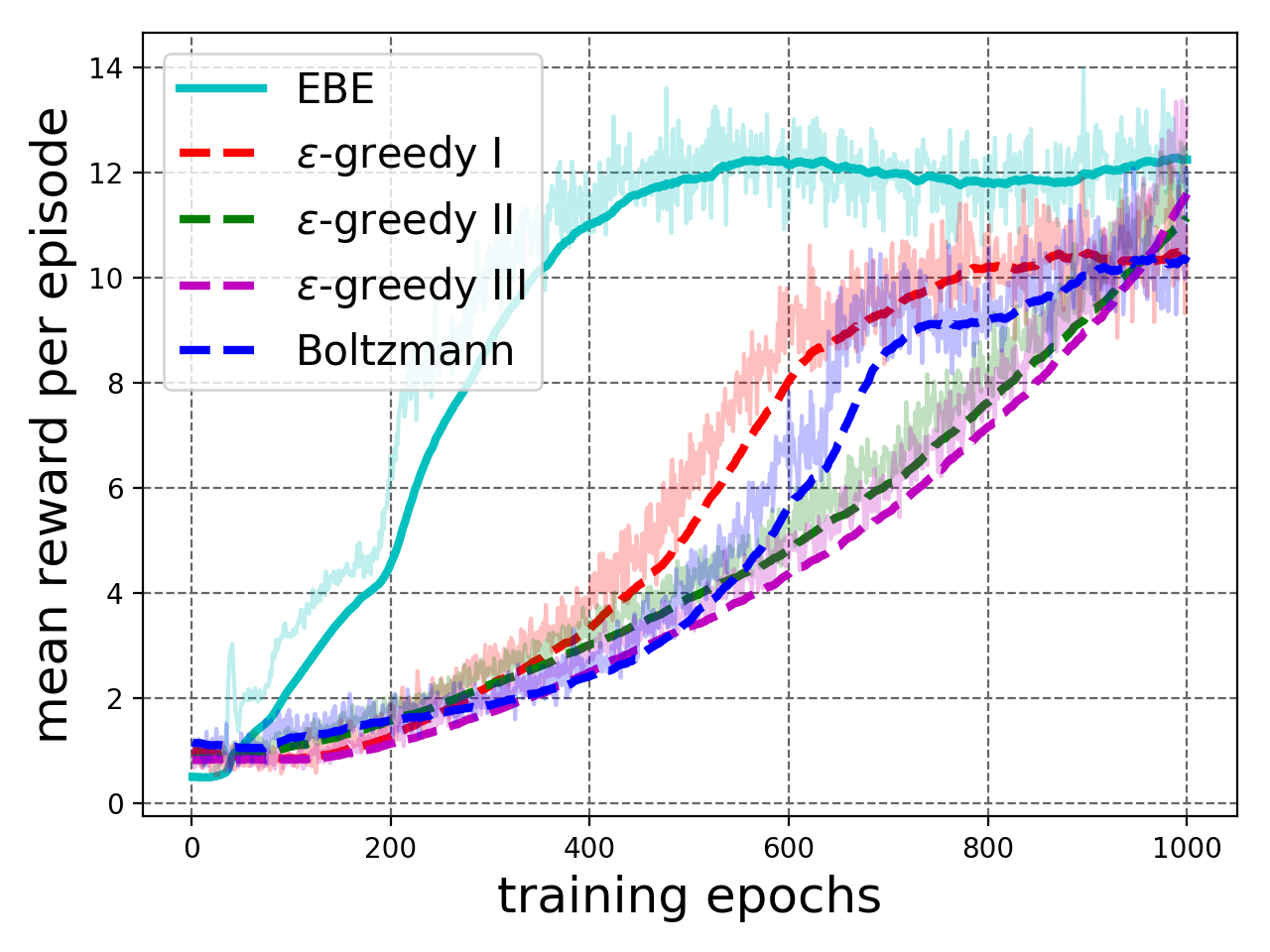}
    }
    \subfigure[]
    {
        \includegraphics[scale=.25]{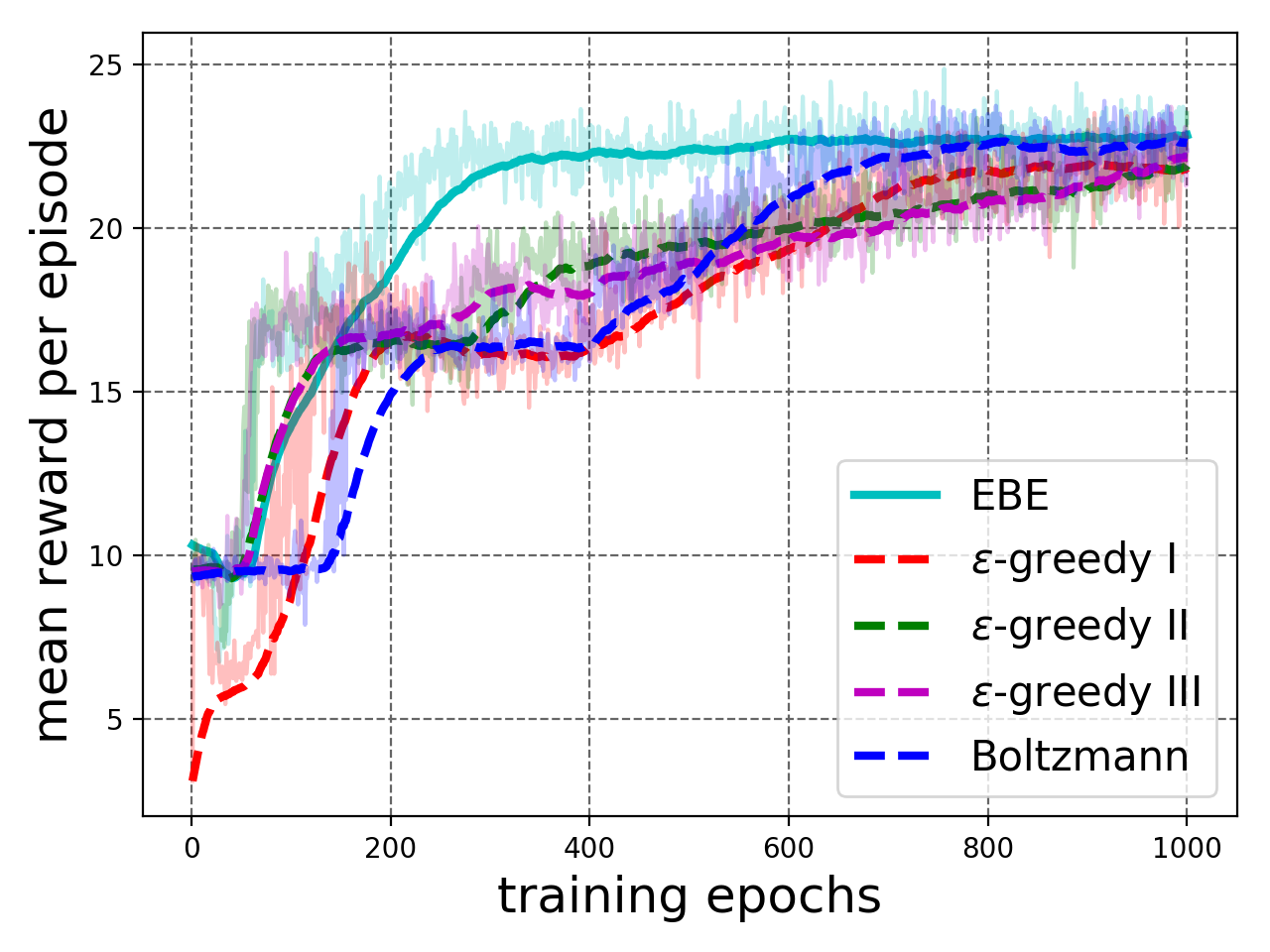}
    }
    \subfigure[]
    {
        \includegraphics[scale=.25]{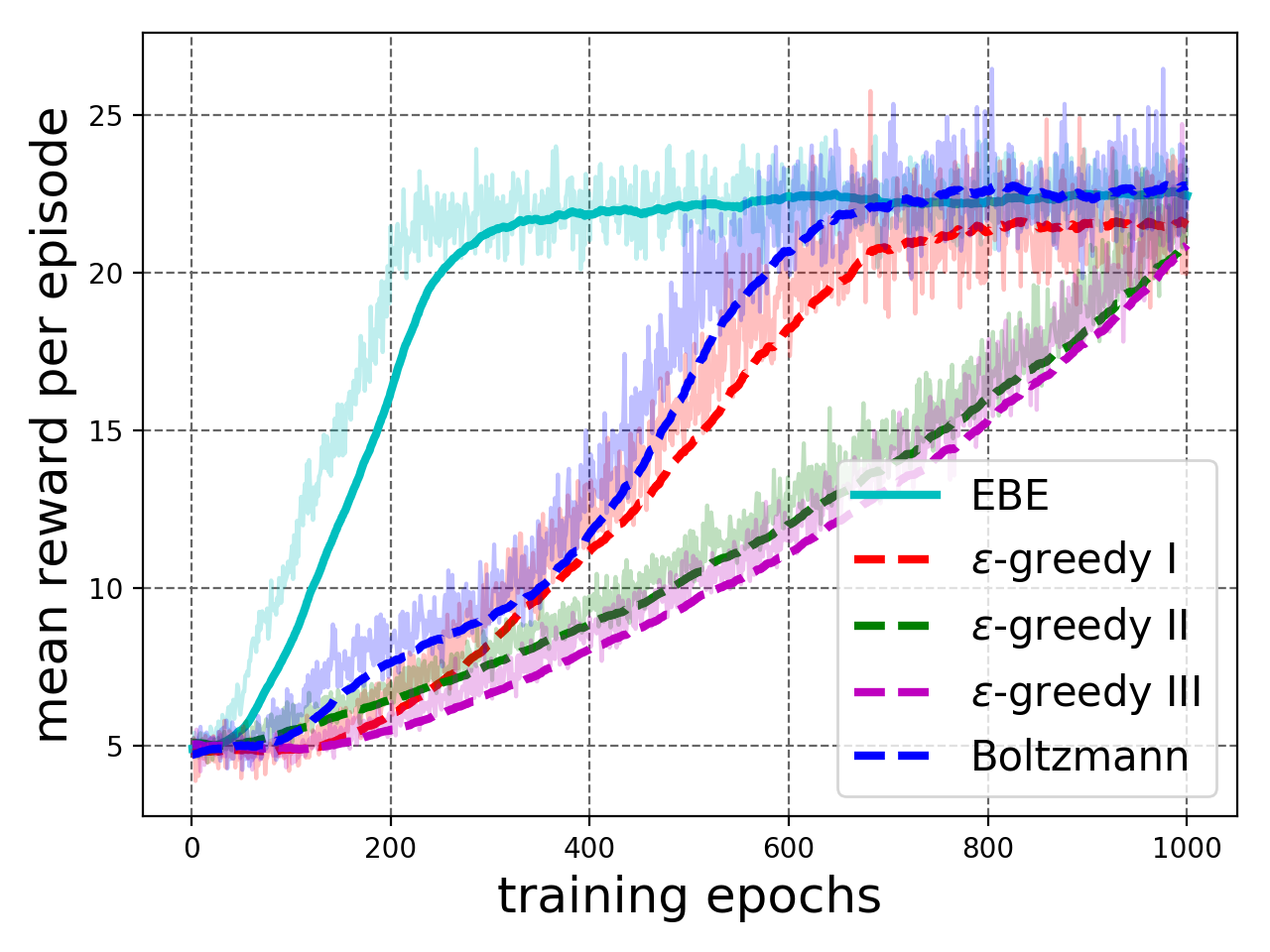}
    }
    \caption
    {
        (a) plots mean test reward of 100 test episodes played after each training epoch while (b) plots mean training reward of all training episode per epoch for for game \textit{DTC}. (c) plots mean test reward of 100 test episodes played after each training epoch while (d) plots mean training reward of all training episode per epoch for for game \textit{DTL}. We compare EBE with $\epsilon$-greedy and Boltzmann exploration strategies. Plots show smoothed data while unsmoothed data is ghosted in the background. Smoothing method is adopted from \cite{tensorboard} with weight 0.975.
    }
    \label{fig:dtc_results_dtl_results}
\end{figure*}

In this environment, the agent is tasked to shoot at attacking monsters spawned around it in a circle. It can only rotate about its position. The agent is provided with 26 ammo and it gets a reward of $1$ point for each kill and $-1$ for getting killed itself. The episode ends when the agent is dead or 2100 steps (60 seconds) have passed. The agent observes the state using raw frames and can either attack, turn left and turn right in a state. An episode is considered successful if the agent kills at least 11 monsters before being dead itself, i.e. scores at least 10 points.

We compare EBE with baselines detailed in Table \ref{tab:variants-of-egreedy}.
Details about the experimental setup are given in Appendix \ref{appendix-vizdoom}.

The experimental results are shown in Figure \ref{fig:dtc_results_dtl_results}. We see in Figure \ref{fig:dtc_results_dtl_results}(a) that the agent trained with EBE exploration attains the maximum mean test reward per episode after about 60\% of training epochs as compared to the other exploration strategies. Moreover, Figure \ref{fig:dtc_results_dtl_results}(b) shows deep exploration, defined in Section \ref{section:deep_exploration}, where EBE was able to perform high reward training episodes early on in the training process. This result shows effectiveness of EBE on high-dimensional RL task that enables effective exploration without having to tune any hyperparameter.

\subsubsection{Defend the Line (DTL)}
This environment is similar to \textit{DTC} except that the agent placed is on one side of the room and monsters are spawning on the opposite wall. The agent is rewarded one point for each kill and penalized one point for being dead. Here, the agent is provided with unlimited ammunition and limited health that decreases with each attack the agent takes from the monsters. The agent observes raw frames and can attack, turn left or turn right in a state. The episode ends when the agent is dead or episode times out with 2100 steps (60 seconds). The goal is to kill at least 16 monsters before the agent dies, i.e. to obtain at least 15 points in one episode.
EBE is compared to the same baselines as considered in Section \ref{subsubsection:dtc}, see Table \ref{tab:variants-of-egreedy}. Details about the experimental setup are given in Appendix \ref{appendix-vizdoom}.

The experimental results are shown in Figure \ref{fig:dtc_results_dtl_results}. Figure \ref{fig:dtc_results_dtl_results}(c) shows that agent trained with EBE exploration attains the maximum mean test reward after about 30\% of training epochs as compared to other exploration strategies. Moreover, Figure \ref{fig:dtc_results_dtl_results}(d) shows deep exploration, defined in Section \ref{section:deep_exploration}, where EBE was able to perform high reward training episodes early on in the training process.

\subsection{Comparison of EBE with Count-Based Exploration}
\label{sec:count-based}
Some of the classic and theoretically-justified exploration methods are based on counting state-action visitations and turning this count into a bonus reward to guide exploration. In the bandit setting, the widely-known Upper-Confidence-Bound (UCB) \cite{ucb} chooses the action $a_t$ that maximizes $\hat{r}(a_t)+\sqrt{\nicefrac{2\log{t}}{N(a_t)}}$ where $\hat{r}(a_t)$ is the estimated reward of executing $a_t$ and $N(a_t)$ is the number of times the action $a_t$ was previously chosen. Similar algorithms for MDP setting select action $a_t$ at time $t$ that maximizes $\Tilde{c}(s_t,a_t) = Q(s_t,a_t) + \mathcal{B}(N(s_t,a_t))$ where $N(s_t,a_t)$ is the number of visitations of $(s_t,a_t)$. Here, $\mathcal{B}(N(s_t,a_t))$ is the exploration bonus that decreases with the increase in $N(s_t,a_t)$. Model Based Interval Estimation-Exploration Bonus (MBIE-EB)\cite{MBIE-EB} proposed $\mathcal{B}(N(s_t,a_t)) = \nicefrac{\beta}{\sqrt{N(s_t,a_t)}}$ with $\beta$ constant. For MDPs, we can get $\mathcal{B}(N(s_t,a_t)) = \sqrt{\nicefrac{2\log{t}}{N(s_t,a_t)}}$.

\begin{figure*}[t]
    \centering
    \subfigure[]
    {
        \includegraphics[scale=.255]{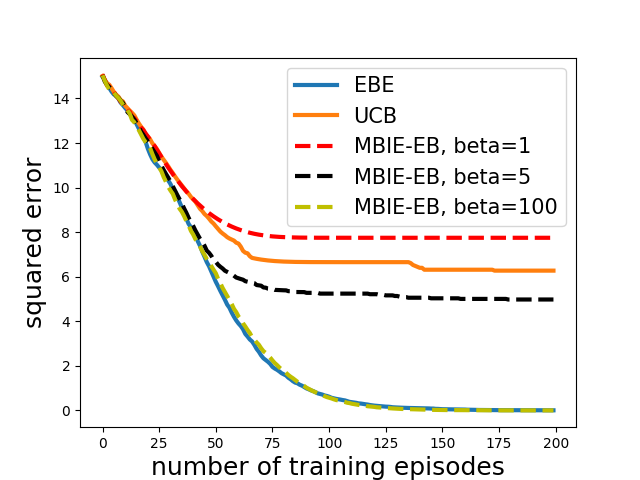}
    }
    \subfigure[]
    {
        \includegraphics[scale=.255]{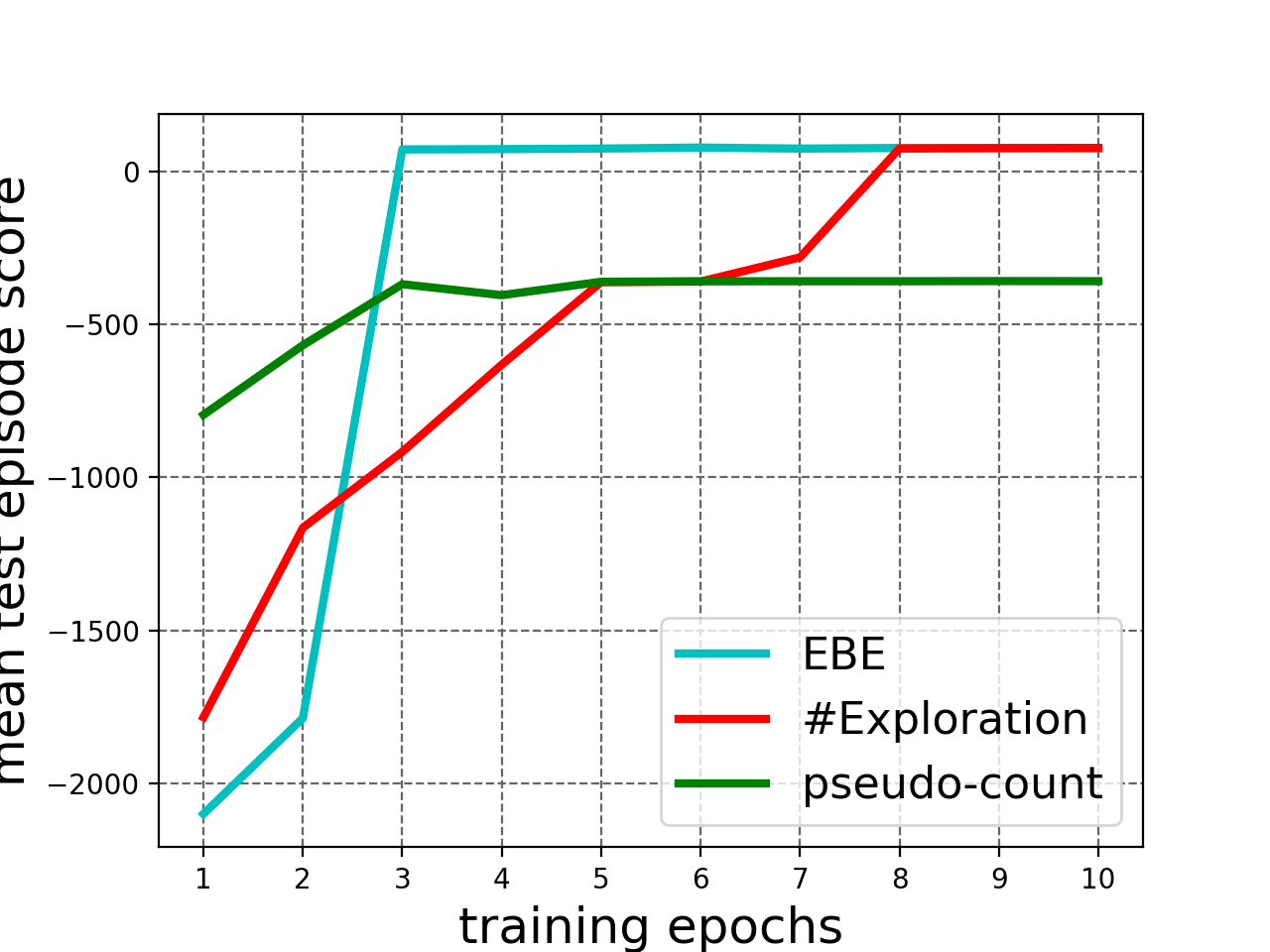}
    }
    \subfigure[]
    {
        \includegraphics[scale=.245]{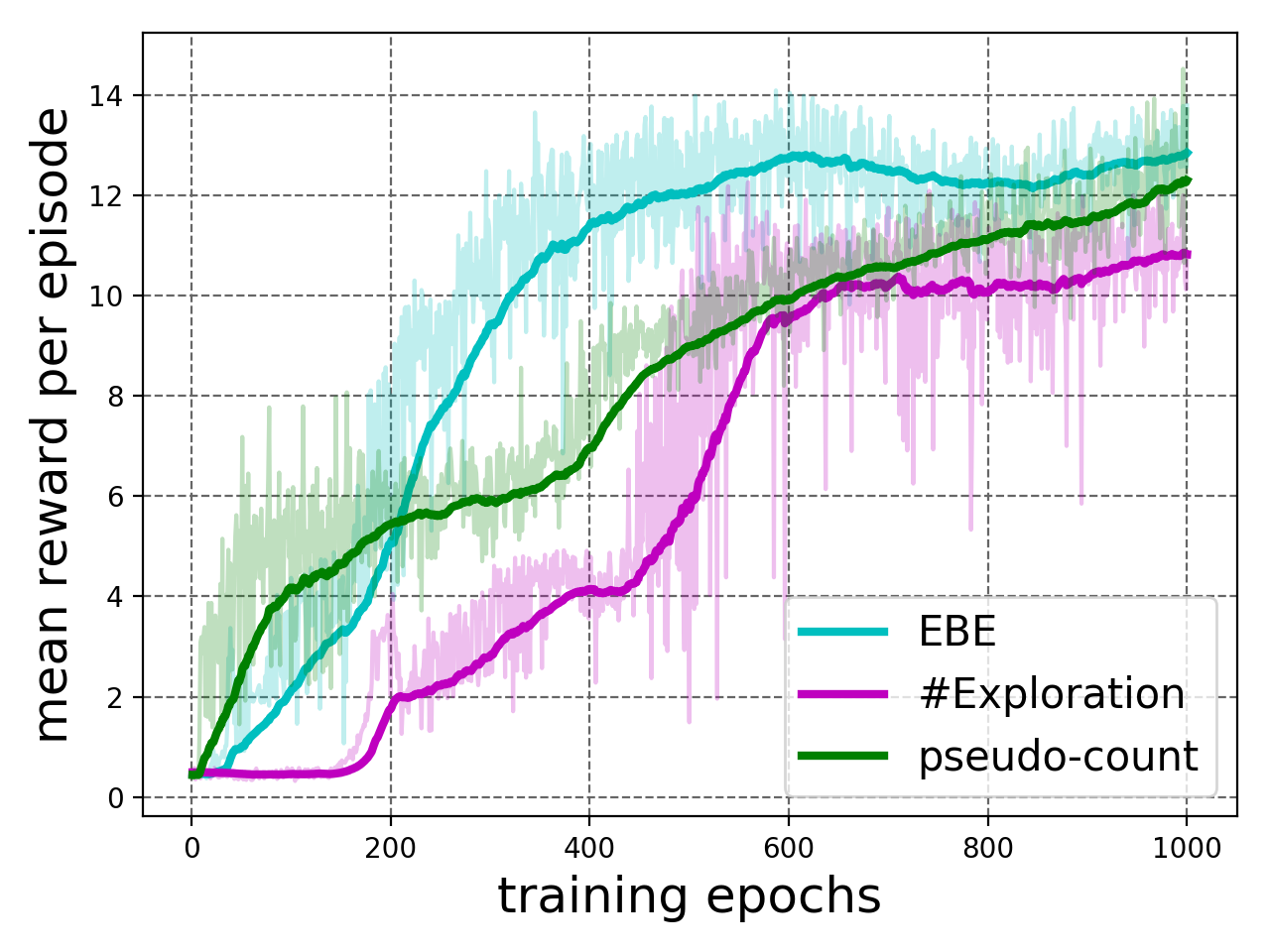}
    }
    \subfigure[]
    {
        \includegraphics[scale=.245]{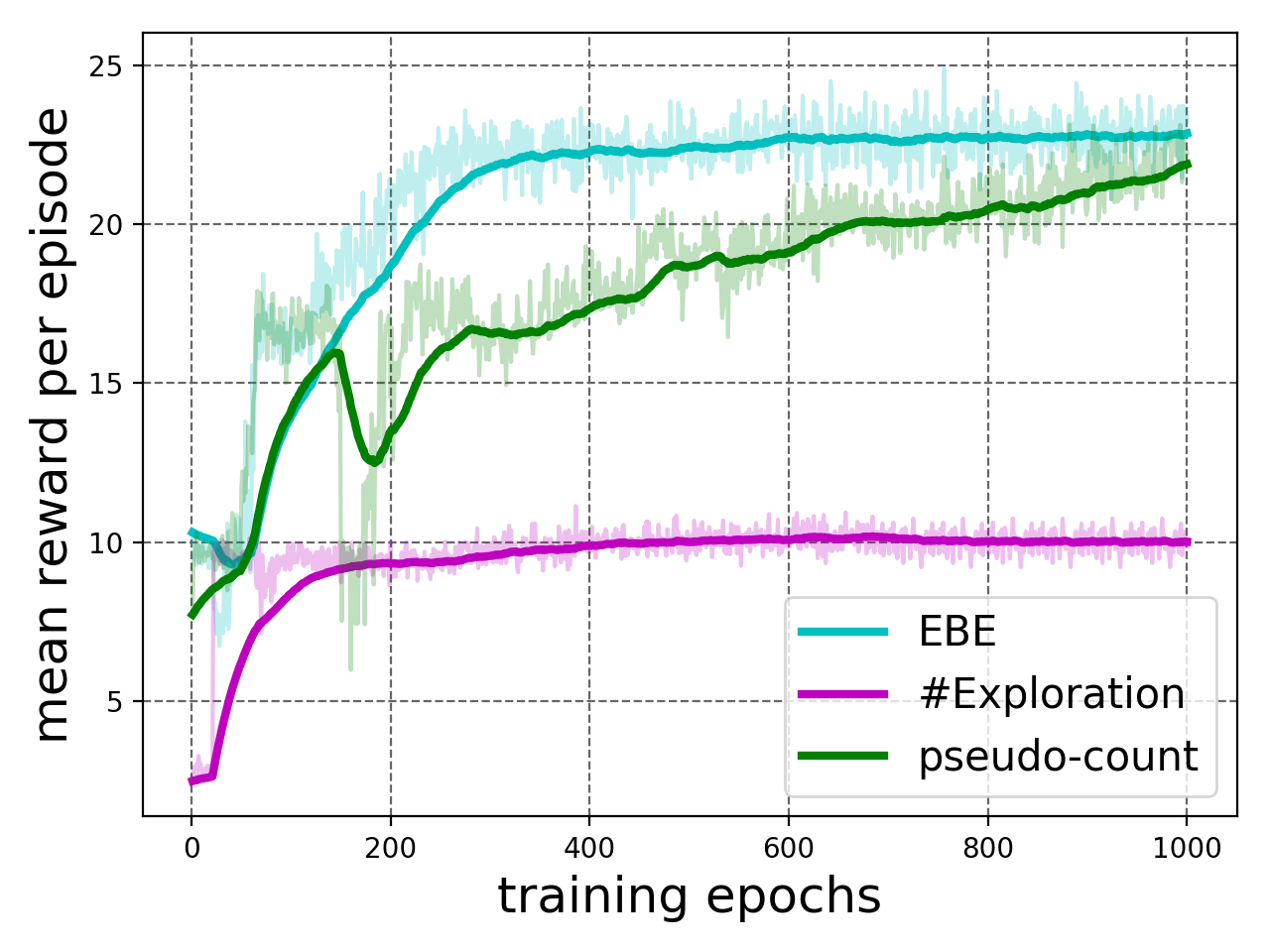}
    }
    \caption
    {
    (a) Comparison of EBE with UCB and MBIE-EB on linear environment. (b) Comparison of EBE with $\#$Exploration and pseudo-count based exploration on VizDoom game Seek and Destroy.
    Comparison of EBE with \textit{\#Exploration} and \textit{pseudo-count} based exploration methods on VizDoom games (c) DTC and (d) DTL.
    }
    \label{fig:count_linear_env_and_snd_dtc_and_dtl_count_results}
    \vspace{-12.5pt}
\end{figure*}

We compare EBE with UCB and MBIE-EB on linear MDP environment considered in Section \ref{subsection:linear} under the same experiment settings. As shown in Figure \ref{fig:count_linear_env_and_snd_dtc_and_dtl_count_results}(a), EBE performs better than UCB in terms of convergence. The performance of MBIE-EB improves as the value of $\beta$ is increased and with $\beta=100$, the performance of MBIE-EB becomes comparable to EBE.

MBIE-EB, UCB and related algorithms assume that the MDP is solved analytically at each timestep, which is only practical for small finite state spaces. Therefore, counting-based methods cannot be extended to high-dimensional, continuous state spaces where states are rarely visited more than once. \cite{intrinsic_motivation_and_count_based} allows generalization of count-based exploration algorithms to the non-tabular case by deriving \textit{pseudo-counts} from arbitrary density models over the state space. $\#$Exploration algorithm \cite{hash_exploration} uses hashing to discretize the high-dimensional state space which visitation counts using a hash table.

We compare EBE with \textit{pseudo-count} based exploration algorithm\cite{intrinsic_motivation_and_count_based} and \textit{\#Exploration}\cite{hash_exploration}. See Appendix \ref{appendix-count-based} for implementation details of these baselines. Figure \ref{fig:count_linear_env_and_snd_dtc_and_dtl_count_results}(b) shows the results for VizDoom game Seek and Destroy. EBE and \#Exploration are able to learn solving the task with EBE learning much earlier while pseudo-count algorithm failed to solve the task. Similarly, Figure \ref{fig:count_linear_env_and_snd_dtc_and_dtl_count_results}(c) and Figure \ref{fig:count_linear_env_and_snd_dtc_and_dtl_count_results}(d) show comparison results for DTC and DTL, respectively. For both games DTC and DTL, EBE depicts efficient exploration by learning to solve the tasks with higher rewards much earlier than the baselines. However, \#Exploration strategy settles at a much lower score for both the games. The following table provides the wall time (in hours) averaged across five runs for DTL and DTC. 
\begin{table}[H]
\vskip -0.1in
\begin{center}
\begin{small}
\begin{tabular}{l p{1 cm} p{1 cm} p{1 cm} p{1 cm} p{1 cm}}
\toprule
game &  EBE & $\epsilon$-greedy & Boltz. & \#Expl. & pseudo-count  \\
\midrule
DTC & 39 & 38 & 42.5 & 64 & 51.5 \\
DTL & 40 & 38 & 44 & 61.5 & 52 \\
\bottomrule
\end{tabular}
\end{small}
\end{center}
\vskip -0.25in
\label{tab:wall_time}
\end{table}
$\epsilon$-greedy exploration is the most efficient in terms of wall time, followed by EBE. \#Exploration incurs needs training time due to online training of autoencoder used for hash codes \cite{hash_exploration} as well.


\subsection{Swing-up Control of Rotary Inverted Pendulum}
\begin{figure}[h]
    \centering
    \subfigure[]{\includegraphics[scale=0.4]{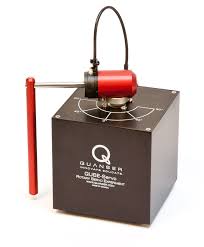}}\quad
    \subfigure[]{\includegraphics[scale=0.15]{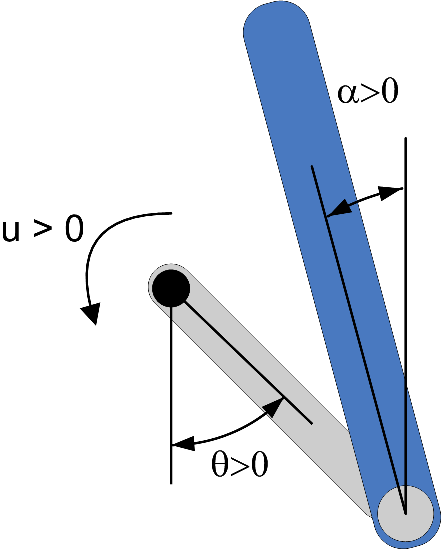}}
    \subfigure[]{\includegraphics[scale=0.3]{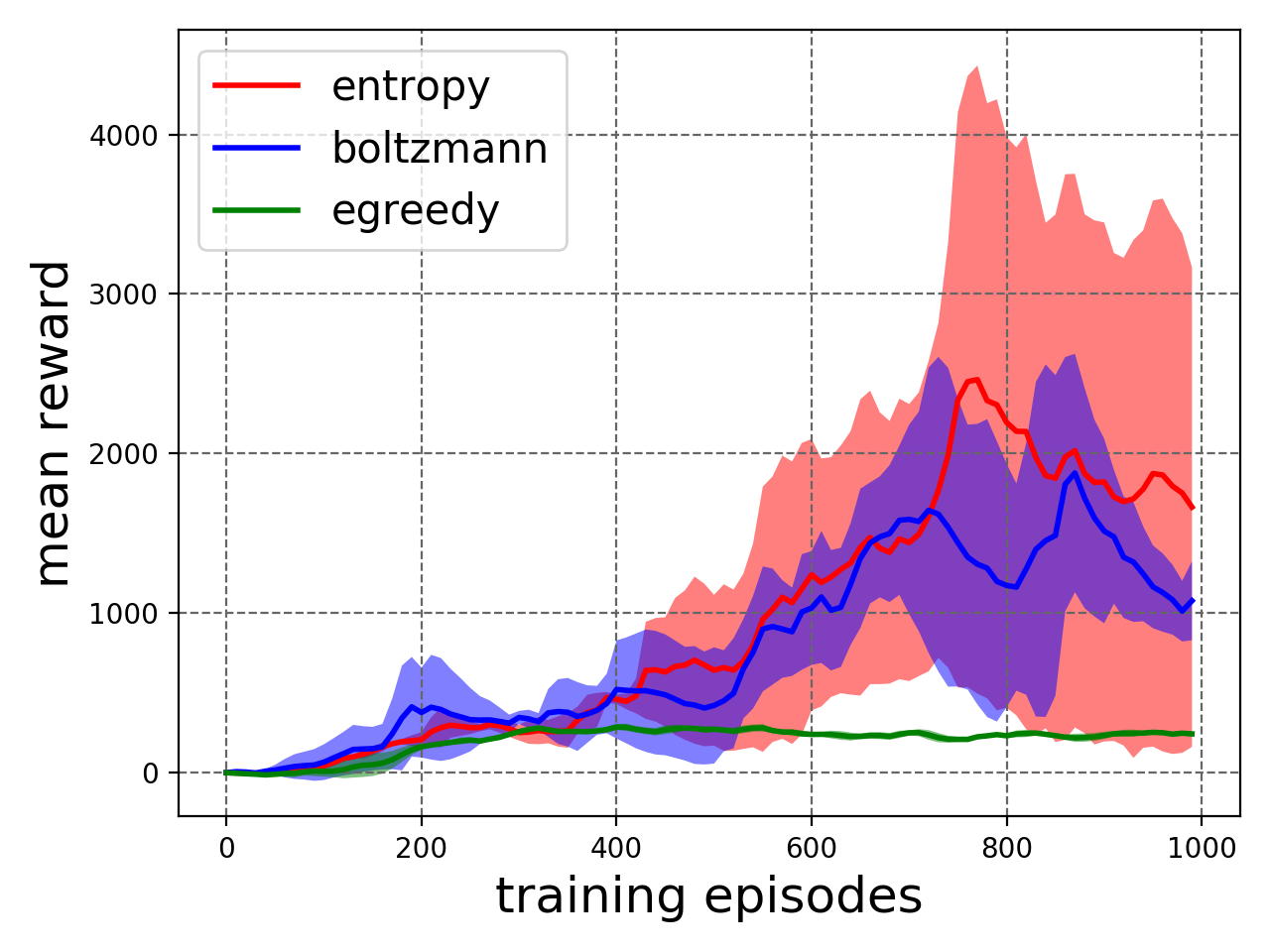}}
    \caption{(a) Quanser Qube. (b) Free Body Diagram of Rotary Inverted Pendulum. $\alpha$ is inverted pendulum angle with respect to its upright position, i.e. $\alpha=0$ when pendulum is upright, $\theta$ is CCW angle of horizontal arm and $u$ is the voltage applied to the motor. Note that $+u$ rotates the horizontal arm in CCW direction. (c) Plot of mean reward of five independent runs for swing-up control of inverted pendulum.}
    \label{fig:quanser_and_free_body}
    \vskip -0.15in
\end{figure}

We compare EBE with Boltzmann and $e$-greedy exploration to perform the swing-up control of a rotary inverted pendulum. We use Quanser Qube \cite{quanser} as our experimental platform as shown in Figure \ref{fig:quanser_and_free_body}(a). We use deep $Q$-learning with prioritized experience replay \cite{per} to learn a policy to swing the pendulum up to its upright position and use a PD controller to balance it in its upright position. The state observation consists of angle $\theta$ of horizontal arm, angle $\alpha$ of pendulum with respect to its upright position and angular velocities $\Dot{\theta}$ and $\Dot{\alpha}$. Angles $\theta$ and $\alpha$ are shown in Figure \ref{fig:quanser_and_free_body}(b). The reward function used for agent learning is given as
\[
    R =
    \begin{cases} 
      L  & \text{when $|\alpha| \leq 25\degree $,} \\
         -L  & \text{when $|\theta| \geq 90\degree $,} \\
        \frac{\pi - 0.8|\theta| - 0.2|\alpha|}{\pi} & \text{otherwise,}
  \end{cases}
\]
where $L = 100$. The RL agent swings the pendulum upright and we switch to PD controller when $|\alpha| \leq 20\degree$. We perform training on the simulator provided by \cite{simulator} and transfer the learnt policies to real system without any fine tuning.

We compare EBE exploration to the following two baselines: $\epsilon-$greedy exploration and Boltzmann exploration. For $\epsilon-$greedy exploration, we use $\epsilon_{i+1}=\epsilon_i - 5e^{-6}$, where $\epsilon_0=1.0$, and for Boltzmann exploration we use temperature value of $1.0$. For training setup details, see Appendix \ref{appendix-pendulum}. The results are shown in Figure \ref{fig:quanser_and_free_body}(c) which plots the mean reward of five independent runs. We see that EBE learns higher reward episodes than any of the baselines. We also visualize the quality of learnt policies in our supplementary video provided at \url{https://youtu.be/nJggIjjzKic}.

\section{Related Work and Discussion}
Existing entropy-based exploration strategies can be broadly divided into two categories \cite{diversity}: entropy regularization \cite{DDP} for RL and maximum entropy principle for RL. Entropy regularization methods, such as \cite{relativeEP, TRPO, PPO}, attempt to alleviate the problem of premature convergence in policy search by imposing information-theoretic constraints on the learning process. \cite{MDPs} shows that entropy regularization yields better optimization properties. Maximum entropy principle methods for RL aim to encourage exploration by optimizing a maximum entropy objective. Authors in \cite{deepEnergyBasedPolicies, softAC, diversityIsAllYouNeed, vime, Ziebart10modelingpurposeful} simply augment the conventional RL objective with the entropy of the policy. \cite{Todorov2007,Todorov2009CompositionalityOO} used the maximum entropy principle to make MDPs linearly solvable while \cite{Fox} employed the maximum entropy principle to incorporate prior knowledge into the RL setting.

Our proposed method belongs to the class of methods that use quantification of uncertainty for exploration. \cite{Still2012} maximizes the information that the most recent state-action pair carries about the future, while \cite{DBLP:journals/corr/HouthooftCDSTA16} maximizes the information gain about the agent's belief of the environment dynamics. Using information gain for exploration can be traced to \cite{infor1} and has been further explored in \cite{Still2012,infor2,infor3}.

Another class of exploration methods, such as \cite{Schmidhuber:1991,Stadie2015IncentivizingEI,DBLP:journals/corr/AchiamS17,pathakICMl17curiosity}, focusses on predicting the environment dynamics where prediction error is used as a basis of exploration. These methods, however, tend to suffer from the \textit{noisy TV} problem \cite{pathakICMl17curiosity} in stochastic and partially-observable MDPs.

Practical reinforcement learning algorithms often utilize simple exploration heuristics, such as $\epsilon$-greedy exploration and Boltzmann exploration \cite{suttonAndBarto}. These methods, however, exhibit a random exploratory behavior, which can lead to exponential regret even in the case of simple MDPs. 

Our proposed method differs from the existing entropy exploration methods for RL in the sense that unlike imposing entropy constraints on old and new policies in entropy regularization methods, we use the entropy to dictate the need for exploration in a state. Soft actor-critic (SAC) \cite{softAC} augments the utility objective with the entropy of policy to motivate exploration while DIAYN \cite{diversityIsAllYouNeed} unsupervisedly learns skills optimizing maximum entropy objective alone. VIME \cite{vime} learns dynamics model $\mathcal{P}(s_{t+1}|s_t, a_t)$ using a Bayesian neural network and optimizes an objective based on an intrinsic reward obtained from the information gain of $\mathcal{P}$ and some utility extrinsic reward.
On the other hand, we focus on optimizing objective based on task utility alone unlike maximum entropy principle methods where the optimizable objective is altered to improve the exploratory behavior of the agent \cite{deepEnergyBasedPolicies, softAC, diversityIsAllYouNeed}. Also, unlike VIME \cite{vime}, we do not need to learn the dynamics model $\mathcal{P}$. This allows the agent to exhibit efficient exploration while optimizing the task utility objective only, thus, maximizing the performance.

\section{Conclusion}
We have introduced a simple-to-implement yet effective exploration strategy that intelligently explores the state space based on agent's learning. We show that the entropy of state-dependent action values can be used to estimate agent's learning for a set of states.
Based on agent's learning, the proposed entropy-based exploration (EBE) is able to decipher the need for exploration in a state, thus, exploring more the unexplored region of state space. This results into what we call deep exploration which is confirmed by multiple experiments on diverse platforms. As shown by the experiments, EBE results into faster and better learning on tabular and high-dimensional state space platforms without having to tune any hyperparameter.

\section{ACKNOWLEDGMENT}
This research has been in part supported by the ICT R\&D program of MSIP/IITP [2016-0-00563, Research on Adaptive Machine Learning Technology Development for Intelligent Autonomous Digital Companion].


\bibliographystyle{ieeetr}
\bibliography{references}

\begin{thebibliography}{10}

\bibitem{suttonAndBarto}
R.~S. Sutton and A.~G. Barto, {\em Reinforcement learning - an introduction}.
\newblock Adaptive computation and machine learning, {MIT} Press, 1998.

\bibitem{humanlevel}
V.~Mnih, K.~Kavukcuoglu, D.~Silver, A.~A. Rusu, J.~Veness, M.~G. Bellemare,
  A.~Graves, M.~Riedmiller, A.~K. Fidjeland, G.~Ostrovski, S.~Petersen,
  C.~Beattie, A.~Sadik, I.~Antonoglou, H.~King, D.~Kumaran, D.~Wierstra,
  S.~Legg, and D.~Hassabis, ``Human-level control through deep reinforcement
  learning,'' {\em Nature}, vol.~518, pp.~529--533, Feb. 2015.

\bibitem{osband_deep_exploration}
I.~Osband, C.~Blundell, A.~Pritzel, and B.~Van~Roy, ``Deep exploration via
  bootstrapped dqn,'' in {\em Advances in Neural Information Processing Systems
  29} (D.~D. Lee, M.~Sugiyama, U.~V. Luxburg, I.~Guyon, and R.~Garnett, eds.),
  pp.~4026--4034, Curran Associates, Inc., 2016.

\bibitem{dueling_networks}
Z.~Wang, T.~Schaul, M.~Hessel, H.~Van~Hasselt, M.~Lanctot, and N.~De~Freitas,
  ``Dueling network architectures for deep reinforcement learning,'' in {\em
  Proceedings of the 33rd International Conference on International Conference
  on Machine Learning - Volume 48}, ICML'16, pp.~1995--2003, JMLR.org, 2016.

\bibitem{kaiming_initialization}
K.~He, X.~Zhang, S.~Ren, and J.~Sun, ``Delving deep into rectifiers: Surpassing
  human-level performance on imagenet classification,'' {\em CoRR},
  vol.~abs/1502.01852, 2015.

\bibitem{vizdoom}
M.~Kempka, M.~Wydmuch, G.~Runc, J.~Toczek, and W.~Jaskowski, ``Vizdoom: {A}
  doom-based {AI} research platform for visual reinforcement learning,'' {\em
  CoRR}, vol.~abs/1605.02097, 2016.

\bibitem{tensorboard}
W.~Chargin and D.~Mor\'e, ``Tensorboard smoothing implementation.''
  \url{https://github.com/tensorflow/tensorboard/blob/f801ebf1f9fbfe2baee1ddd65714d0bccc640fb1/tensorboard/plugins/scalar/vz_line_chart/vz-line-chart.ts#L704},
  2015.

\bibitem{ucb}
T.~L. Lai and H.~Robbins, ``Asymptotically efficient adaptive allocation
  rules,'' {\em Advances in Applied Mathematics}, vol.~6(1):4-22, 1985.

\bibitem{MBIE-EB}
T.~L. Lai and H.~Robbins, ``An analysis of model-based interval estimation for
  markov decision processes,'' {\em Journal of Computer and System Sciences},
  vol.~74(8):1309-1331, 2008.

\bibitem{intrinsic_motivation_and_count_based}
M.~G. Bellemare, S.~Srinivasan, G.~Ostrovski, T.~Schaul, D.~Saxton, and
  R.~Munos, ``Unifying count-based exploration and intrinsic motivation,'' {\em
  CoRR}, vol.~abs/1606.01868, 2016.

\bibitem{hash_exploration}
H.~Tang, R.~Houthooft, D.~Foote, A.~Stooke, X.~Chen, Y.~Duan, J.~Schulman,
  F.~D. Turck, and P.~Abbeel, ``{\#}exploration: {A} study of count-based
  exploration for deep reinforcement learning,'' {\em CoRR},
  vol.~abs/1611.04717, 2016.

\bibitem{quanser}
Quanser, ``{Quanser Qube Servo 2}.''
  \url{https://www.quanser.com/products/qube-servo-2/}, 2020.

\bibitem{per}
T.~Schaul, J.~Quan, I.~Antonoglou, and D.~Silver, ``Prioritized experience
  replay,'' 2015.
\newblock cite arxiv:1511.05952Comment: Published at ICLR 2016.

\bibitem{simulator}
K.~{Polzounov}, R.~{Sundar}, and L.~{Redden}, ``{Blue River Controls: A toolkit
  for Reinforcement Learning Control Systems on Hardware}.'' Accepted at the
  Workshop on Deep Reinforcement Learning at the 33rd Conference on Neural
  Information Processing Systems (NeurIPS 2019), Vancouver, Canada., 2019.

\bibitem{diversity}
Z.-W. Hong, T.-Y. Shann, S.-Y. Su, Y.-H. Chang, T.-J. Fu, and C.-Y. Lee,
  ``Diversity-driven exploration strategy for deep reinforcement learning,'' in
  {\em Advances in Neural Information Processing Systems 31} (S.~Bengio,
  H.~Wallach, H.~Larochelle, K.~Grauman, N.~Cesa-Bianchi, and R.~Garnett,
  eds.), pp.~10510--10521, Curran Associates, Inc., 2018.

\bibitem{DDP}
M.~G. Azar and H.~J. Kappen, ``Dynamic policy programming,'' {\em CoRR},
  vol.~abs/1004.2027, 2010.

\bibitem{relativeEP}
J.~Peters, K.~M{\"u}lling, and Y.~Altun, ``Relative entropy policy search,'' in
  {\em AAAI 2010}, 2010.

\bibitem{TRPO}
J.~Schulman, S.~Levine, P.~Abbeel, M.~Jordan, and P.~Moritz, ``Trust region
  policy optimization,'' in {\em Proceedings of the 32nd International
  Conference on Machine Learning} (F.~Bach and D.~Blei, eds.), vol.~37 of {\em
  Proceedings of Machine Learning Research}, (Lille, France), pp.~1889--1897,
  PMLR, 07--09 Jul 2015.

\bibitem{PPO}
J.~Schulman, F.~Wolski, P.~Dhariwal, A.~Radford, and O.~Klimov, ``Proximal
  policy optimization algorithms,'' {\em CoRR}, vol.~abs/1707.06347, 2017.

\bibitem{MDPs}
G.~Neu, A.~Jonsson, and V.~G{\'{o}}mez, ``A unified view of entropy-regularized
  markov decision processes,'' {\em CoRR}, vol.~abs/1705.07798, 2017.

\bibitem{deepEnergyBasedPolicies}
T.~Haarnoja, H.~Tang, P.~Abbeel, and S.~Levine, ``Reinforcement learning with
  deep energy-based policies,'' in {\em ICML}, 2017.

\bibitem{softAC}
T.~Haarnoja, A.~Zhou, P.~Abbeel, and S.~Levine, ``Soft actor-critic: Off-policy
  maximum entropy deep reinforcement learning with a stochastic actor,'' {\em
  CoRR}, vol.~abs/1801.01290, 2018.

\bibitem{diversityIsAllYouNeed}
B.~Eysenbach, A.~Gupta, J.~Ibarz, and S.~Levine, ``Diversity is all you need:
  Learning skills without a reward function,'' {\em CoRR}, vol.~abs/1802.06070,
  2018.

\bibitem{vime}
R.~Houthooft, X.~Chen, Y.~Duan, J.~Schulman, F.~Turck, and P.~Abbeel, ``Vime:
  Variational information maximizing exploration,'' in {\em NIPS}, 2016.

\bibitem{Ziebart10modelingpurposeful}
B.~D. Ziebart and M.~Hebert, ``Modeling purposeful adaptive behavior with the
  principle of maximum causal entropy,'' 2010.

\bibitem{Todorov2007}
E.~Todorov, ``Linearly-solvable markov decision problems,'' in {\em Advances in
  Neural Information Processing Systems 19} (B.~Sch\"{o}lkopf, J.~C. Platt, and
  T.~Hoffman, eds.), pp.~1369--1376, MIT Press, 2007.

\bibitem{Todorov2009CompositionalityOO}
E.~Todorov, ``Compositionality of optimal control laws,'' in {\em NIPS}, 2009.

\bibitem{Fox}
R.~Fox, A.~Pakman, and N.~Tishby, ``Taming the noise in reinforcement learning
  via soft updates,'' in {\em Proceedings of the Thirty-Second Conference on
  Uncertainty in Artificial Intelligence}, UAI'16, (Arlington, Virginia, United
  States), pp.~202--211, AUAI Press, 2016.

\bibitem{Still2012}
S.~Still and D.~Precup, ``An information-theoretic approach to curiosity-driven
  reinforcement learning,'' {\em Theory in Biosciences}, vol.~131,
  pp.~139--148, Sep 2012.

\bibitem{DBLP:journals/corr/HouthooftCDSTA16}
R.~Houthooft, X.~Chen, Y.~Duan, J.~Schulman, F.~D. Turck, and P.~Abbeel,
  ``Curiosity-driven exploration in deep reinforcement learning via bayesian
  neural networks,'' {\em CoRR}, vol.~abs/1605.09674, 2016.

\bibitem{infor1}
J.~Storck, S.~Hochreiter, and J.~Schmidhuber, ``Reinforcement driven
  information acquisition in non-deterministic environments,'' 1995.

\bibitem{infor2}
Y.~Sun, F.~J. Gomez, and J.~Schmidhuber, ``Planning to be surprised: Optimal
  bayesian exploration in dynamic environments,'' {\em CoRR},
  vol.~abs/1103.5708, 2011.

\bibitem{infor3}
D.~Y. Little and F.~T. Sommer, ``Learning and exploration in action-perception
  loops,'' in {\em Front. Neural Circuits}, 2013.

\bibitem{Schmidhuber:1991}
J.~Schmidhuber, ``A possibility for implementing curiosity and boredom in
  model-building neural controllers,'' in {\em Proceedings of the First
  International Conference on Simulation of Adaptive Behavior on From Animals
  to Animats}, (Cambridge, MA, USA), pp.~222--227, MIT Press, 1990.

\bibitem{Stadie2015IncentivizingEI}
B.~C. Stadie, S.~Levine, and P.~Abbeel, ``Incentivizing exploration in
  reinforcement learning with deep predictive models,'' {\em CoRR},
  vol.~abs/1507.00814, 2015.

\bibitem{DBLP:journals/corr/AchiamS17}
J.~Achiam and S.~Sastry, ``Surprise-based intrinsic motivation for deep
  reinforcement learning,'' {\em CoRR}, vol.~abs/1703.01732, 2017.

\bibitem{pathakICMl17curiosity}
D.~Pathak, P.~Agrawal, A.~A. Efros, and T.~Darrell, ``Curiosity-driven
  exploration by self-supervised prediction,'' in {\em ICML}, 2017.

\bibitem{pixelcnn++}
T.~Salimans, A.~Karpathy, X.~Chen, and D.~P. Kingma, ``Pixelcnn++: Improving
  the pixelcnn with discretized logistic mixture likelihood and other
  modifications,'' {\em CoRR}, vol.~abs/1701.05517, 2017.

\end{thebibliography}




\section*{APPENDIX}
\setcounter{section}{0}
\section{Details about Experimental Setup}
\subsection{Simpler Breakout Game}
\label{appendix-breakout}
DQN is used as learning algorithm for this environment. Two most recent frames, the current one and the previous one, are used as current state information and fed to the $Q$-network to estimate action values for the state. The $Q$-network is a convolutional neural network whose architecture is detailed below:

\begin{table*}[ht]
\caption{$Q$- network architecture for breakout game experiments.}
\label{tab:qnetwork-architecture-breakout}
\begin{center}
\begin{small}
\begin{tabular}{lcccccc} 
\toprule  
Layer & input size & filter size  & stride & no. of filters/neurons & activation & output size \\
\midrule
CONV 1    & $8 \times 5 \times 2$  & $3 \times 3$ & 1 & 32 &  RELU  & $6 \times 3 \times 32$ \\
CONV 2    & $6 \times 3 \times 32$  & $2 \times 2$ & 1 & 64 &  RELU  & $5 \times 2 \times 64$ \\
FC 1      & 640  & - & - & 256 &  RELU  & 256 \\
FC 2      & 256  & - & - & 3 &  LINEAR  & 3 \\
\bottomrule
\end{tabular}
\end{small}
\end{center}
\vskip -0.1in
\end{table*}

We use Adam optimizer with the learning rate of $10^{-4}$. Agents are trained for 3000 episodes with a maximum of 200 steps per epoch and the target network update frequency of 100 steps. We use the minibatch size of 10, discount factor 0.95 and replay memory size of 1000. 

\subsection{VizDoom Experiments}
\label{appendix-vizdoom}
We use DQN as our learning algorithm. Raw gray scale input images of resolution $640 \times 480$ are scaled down to resolution $100 \times 150$. These images are then processed by a deep $Q$-network. The $Q$-network architecture is shown in Table \ref{tab:qnetwork-archotecture}.
\begin{table*}[ht]
\caption{$Q$- network architecture for VizDoom experiments.}
\label{tab:qnetwork-archotecture}
\begin{center}
\begin{small}
\begin{tabular}{l c c c c c c}
\toprule
Layer & input size & filter size  & stride & no. of filters/neurons & activation & output size \\
\midrule
conv 1    & $100 \times 150 \times 1$  & $6 \times 6$ & 3 & 8 &  RELU  & $32 \times 49 \times 8$ \\
conv 2    & $32 \times 49 \times 8$  & $3 \times 3$ & 2 & 8 &  RELU  & $15 \times 24 \times 8$ \\
FC 1      & 2880  & - & - & 128 &  RELU  & 128 \\
FC 2      & 128  & - & - & $|\mathcal{A}|$ &  Linear  & $|\mathcal{A}|$ \\
\bottomrule
\end{tabular}
\end{small}
\end{center}
\vskip -0.1in
\end{table*}

Since the number of available actions for each considered VizDoom game is 3, we have $|\mathcal{A}|=3$ in \ref{tab:qnetwork-archotecture}. We do not use any target network for VizDoom experiments. Also we only use the current frame as state observation. Stochastic Gradient Descent (SGD) is used as optimizer with the learning rate of 0.00025. We use the minibatch size of 64 and the discount factor of 0.99.

For game Seek and Destroy, we use replay memory size of 10000 and train the agent for 10 epochs with 2000 steps per epoch. However, for games defend the center and defend the line, we train the agent for 1000 epochs with 5000 steps per epoch and use the replay memory size of 50000.

Experiments are performed with one NVIDIA TITAN Xp GPU and 12 gigabytes of RAM. It takes approximately 55 hours to complete 1000 epochs for Defend the Line and Defend the Center games while it takes approximately 15 minutes to complete 10 epochs for Seek and Destroy game.

\subsection{Swing-up Control of Rotary Inverted Pendulum Pendulum}
\label{appendix-pendulum}
The action space is discretized to get the following set $\mathcal{A} = \{ -3, -1.8, -0.6, 0.6, 1.8, 3.0 \}$. $Q$-values are approximated using deep neural network having one hidden layer of 128 neurons and relu as activation function and an output layer of $|\mathcal{A}| = 6$ neurons. The network is trained with batch size of 32, discount factor 0.99, learning rate 0.001, replay memory size 50000 and target network update frequency of 1000 steps. The first 10,000 steps are used for initial data collection and no learning is performed during this time. The episode ends when either 2048 steps have passed or $|\theta| \geq 90\degree$. For balance control, we use a PD controller with $k_p^{\theta}=-2.0, k_p^{\alpha}=35.0, k_d^{\theta}=-1.5$ and $k_d^{\alpha}=3.0$. The control law is given as 
\[
u  = k_p^{\theta} \theta + k_p^{\alpha} \alpha +  k_d^{\theta} \Dot{\theta} + k_d^{\alpha} \Dot{\alpha},
\]
where the control input $u$ is further clipped to range $[-3.0, 3.0]$ to prevent any hardware damage.

\section{Comparison of EBE with Count-Based Exploration Methods - Implementation Details}
\label{appendix-count-based}
In this section, we explain the details about implementation of baselines in Section \ref{sec:count-based} of the main paper.
\subsection{Pseudo-Count Based Exploration}
\label{appendix-count-based-pseudo}
We use a gated variant of PixelCNN++ \cite{pixelcnn++} as the density model over the state space that is used to generate the exploration bonus. This bonus is then used to guide the exploration. The pseudo-count is computed as
\[
\hat{N}_n(x) = \sqrt{e^{\frac{0.1 * (PG_n(x))_+}{\sqrt{n+1}}}-1},
\]
where $(PG_n(s))_+ = \max(0, PG_n(s))$ and $PG_n(s):=\log \rho_n^{'}(s) - \log \rho_n(s)$ is the prediction gain and $\rho_n^{'}(s)$ and $\rho_n(s)$ are defined in \cite{intrinsic_motivation_and_count_based}. The agent then selects the action that maximizes $Q(s,a)+\hat{N}_n(s)$ at step $n$.

\subsection{\#Exploration}
\label{appendix-count-hash}
Since our observations consist of raw images, an autoencoder (AE) from \cite{hash_exploration} is used to get the hash codes. See Figure 1 of \cite{hash_exploration} for the AE architecture. The solid block represents the dense sigmoidal binary code layer, after which noise with uniform distribution $\mathcal{U}(-0.3, 0.3)$ is injected to improve AE's capability to reconstruct the distinct state inputs as explained in \cite{hash_exploration}. The code is then rounded to the nearest integer. Matrix $A \in \mathbb{R}^{k \times D}$, with entries drawn i.i.d. from the standard Gaussian distribution $\mathcal{N}(0,1)$ and $k=24$ and $D=512$, is used to project the code to lower dimensional space via SimHash. See Algorithm 2 in \cite{hash_exploration} for more details. The AE is trained using loss function described in equation (3) in \cite{hash_exploration} where we use $\lambda = 1$ and $K=1$. The rest of the implementation details are the same as in Appendix \ref{appendix-vizdoom}.

\end{document}